\documentclass[journal]{IEEEtran}
\usepackage{caption}
\usepackage{multirow}
\usepackage{makecell}
\usepackage{float}
\usepackage{hyperref}
\usepackage{booktabs}
\usepackage{enumitem}
\usepackage{pdfpages}
\usepackage{textcomp}
\usepackage{epsfig,latexsym}
\usepackage{indentfirst}
\usepackage{amsmath}
\usepackage{amssymb}
\usepackage{xcolor}
\usepackage{times}
\usepackage{subfigure}
\usepackage{psfrag}
\usepackage{cite}
\usepackage{amsthm}
\usepackage{bm}
\usepackage{flushend}  
\usepackage{wasysym}
\usepackage{epstopdf}
\usepackage{fancyhdr}
\usepackage{stfloats}
\usepackage{color}
\usepackage[noend]{algpseudocode}
\usepackage{algorithmicx,algorithm}
\usepackage{tabularx}
\usepackage{makecell}
\usepackage{ragged2e}
\usepackage{graphicx}
\newcommand{\xqedhere}[2]{%
\rlap{\hbox to#1{\hfil\llap{\ensuremath{#2}}}}}

\newcommand{\tabincell}[2]{\begin{tabular}{@{}#1@{}}#2\end{tabular}}

\newtheorem{theorem}{\bf{Theorem}}
\newtheorem{assumption}{\bf{Assumption}}
\newtheorem{lemma}{Lemma}

\newtheorem{corollary}{Corollary}
\newtheorem{definition}{\bf{Definition}}

\allowdisplaybreaks
\begin{document}%
\title{Tool-Aided Evolutionary LLM for Generative Policy Toward Efficient  Resource Management in Wireless Federated Learning}

\author{{Chongyang Tan, Ruoqi Wen, Rongpeng Li, Zhifeng Zhao, Ekram Hossain, and Honggang Zhang\vspace{-.75cm}
}
    \thanks{C. Tan, R. Wen, and R. Li are with the College of Information Science and Electronic Engineering, Zhejiang University, Hangzhou 310058, China (email: \{cyatan, wenruoqi, lirongpeng\}@zju.edu.cn).}
    \thanks{Z. Zhao is with Zhejiang Lab, Hangzhou 311121, China, and also with the College of Information Science and Electronic Engineering, Zhejiang University, Hangzhou 310058, China (email: zhaozf@zhejianglab.org).}
    \thanks{Ekram Hossain is with the University of Manitoba, Winnipeg, Canada (email: ekram.hossain@umanitoba.ca).}
    \thanks{Honggang Zhang is with Macau University of Science and Technology, China (email: hgzhang@must.edu.mo).}
}
\maketitle

\begin{abstract}
    Federated Learning (FL) enables distributed model training across edge devices in a privacy-friendly manner. However, its efficiency heavily depends on effective device selection and high-dimensional resource allocation in dynamic and heterogeneous wireless environments. Conventional methods demand a confluence of domain-specific expertise, extensive hyperparameter tuning, and/or heavy interaction cost. This paper proposes a Tool-aided Evolutionary Large Language Model (T-ELLM) framework to generate a qualified policy for device selection in a wireless FL environment. Unlike conventional optimization methods, T-ELLM leverages natural language-based scenario prompts to enhance generalization across varying network conditions. The framework decouples the joint optimization problem mathematically, enabling tractable learning of device selection policies while delegating resource allocation to convex optimization tools. 
    To facilitate the evolutionary process, T-ELLM interacts with a sample-efficient, model-based virtual learning environment that captures the relationship between device selection and learning performance. This developed virtual environment reduces reliance on real-world interactions, thus minimizing communication overhead while refining the LLM-based decision-making policy through group relative policy optimization.
    Theoretical analysis proves that the discrepancy between virtual and real environments is bounded, ensuring the advantage function learned in the virtual environment maintains a provably small deviation from real-world conditions. Experimental results demonstrate that T-ELLM outperforms benchmark methods in energy efficiency and exhibits robust adaptability to environmental changes.
\end{abstract}
\begin{IEEEkeywords}
    Large language model, generative policy, wireless federated learning, resource management, convex optimization, reinforcement learning.
\end{IEEEkeywords}

\vspace{-0.25em}
\section{Introduction}
\IEEEPARstart{N}{ext}-Generation (xG) wireless communication systems are envisioned to support various intelligent applications and services\cite{cui2025overview}, empowered by the exponential growth of wireless edge devices, such as mobile phones and sensors. As a prominent paradigm\cite{mcmahan2017communication,guan2024federated}, Federated Learning (FL) emerges by facilitating collaborative model training across decentralized devices while maintaining data locality in a privacy-friendly manner. Nevertheless, the deployment of FL in wireless networks faces severe challenges, arising from the underlying heterogeneous computation and communication capabilities across devices \cite{10.1145/3625558, 9252927}. Correspondingly, a proper resource management policy for FL, which selects the participating devices and calibrates the utilized communications and computing resources, will need to be developed in dynamic wireless environments \cite{yi_rhfedmtl_2024}. Typically, such a problem results in a high-dimensional optimization problem that can be \emph{partially} solved by heuristic solutions \cite{273723,8761315} or Reinforcement Learning (RL)-based approaches \cite{Zhang_Lin_Zhang_2022,10368103,10.1145/3466752.3480129,9155494,9139873}. However, heuristic solutions often demand a confluence of domain-specific expertise and extensive tuning before adapting to unseen scenarios, while RL-based approaches require heavy interactions with the environment and become sluggish for large system dimensions and changing system dynamics \cite{10078377,10.1145/3054912, Zhang2021}. Therefore, there is a strong incentive to find alternative solutions\cite{LI2024110663}.
\vspace{-0.25em}
\subsection{Related Works}
To implement FL over wireless networks, in each training round, edge devices upload the locally trained updates to a centralized server via wireless links, in exchange for aggregated models. Through several training rounds, the performance of the eventually learned global model is primarily impacted by resource and data heterogeneity. For example, disparities in computational and communication capabilities across devices lead to imbalanced time and energy consumption, while varying dataset sizes and non-Independent and Identically Distributed (non-IID) data distributions can result in biased model updates. Both would degrade the learning effectiveness \cite{10.1145/3625558, 9252927}. Additionally, the deployment of FL encounters constraints due to the energy and time \cite{8761315} budget \cite{10368103,9210812}. For instance,
the authors in \cite{9264742,9187798} formulate the joint learning and communication problem as the minimization of total energy consumption under latency constraints and provide computationally efficient closed-form solutions to optimize critical resources, including bandwidth allocation, CPU frequency, and transmission power. But, full device participation is bluntly assumed in \cite{9264742,9187798}. However, mobilizing the participation of a subset of qualified edge devices in each training round can contribute to improving learning efficiency. For example, biasing client selection towards clients with higher local loss \cite{pmlr-v151-jee-cho22a} proves to yield faster convergence than the random selection in Federated Averaging (FedAvg) \cite{mcmahan2017communication}. Also, a timer can be heuristically set to avoid the participation of straggler devices \cite{8761315}.
In \cite{273723}, a guided participant selection scheme improves FL performance by jointly optimizing system and data utility through an epsilon-greedy algorithm. Moreover, a joint device scheduling and bandwidth allocation policy, which maximizes model accuracy within a fixed training time budget by a hybrid greedy-optimization approach, is proposed in \cite{9207871}, while energy constraints are neglected therein. More importantly, adapting these heuristics to unexpected scenarios, such as tuning hyperparameters to approximate FL performance limits\cite{8761315,9207871} or setting appropriate timers\cite{273723} when system bandwidth or device resources (e.g., maximum transmission power, dataset size) change, typically demands iterative adjustments and domain-specific expertise.

RL provides an alternative methodology to solve wireless FL problems. Through continuously interacting with the environment, RL agents evaluate candidate policies in the wireless FL environment and gradually learn appropriate real-time policies. For example, \cite{9155494} develops a Deep RL (DRL) method to select devices to upload their local model for aggregation, to maximize validation accuracy. Nevertheless, it overlooks energy and time constraints and assumes all devices participate in local training. Similarly, \cite{10.1145/3466752.3480129} adopts a Q-learning-based method to identify near-optimal participating devices. Furthermore, to ameliorate the impact of high dimensionality on single-agent RL with large-scale devices,  \cite{Zhang_Lin_Zhang_2022} introduces a Multi-Agent Reinforcement Learning (MARL) framework that jointly optimizes model accuracy, processing latency, and communication efficiency. On the other hand,  \cite{9139873} improves the energy efficiency of FL by managing the CPU-cycle frequency of mobile devices based on DRL. To further improve the efficiency of FL, the works in \cite{10181138,10381828,9779339} leverage DRL-based approaches to jointly optimize the device selection and bandwidth allocation. However, the above DRL-based methods only optimize partial parameters and rely on extensive training in a given environment. Therefore, the changes in environment or data often imply cumbersome retraining or even architectural redesign. Unfortunately, a real-world wireless scenario constantly evolves, and it inevitably induces substantial engineering cost. Meanwhile, direct interaction with the environment incurs substantial communication overhead. Although imitation learning \cite{10.1145/3054912} can acquire a policy by learning from the optimal trajectories to reduce interaction overhead, there are no such expert systems strictly in this FL scenario.
Besides, the training difficulty, which increases exponentially along with the potentially involved devices and managed resources, limits the practicability \cite{10078377, Zhang2021}.

The remarkable capabilities of transformer-based Large Language Models (LLMs) \cite{Radford2018ImprovingLU, radford2019language, grattafiori2024llama3herdmodels,deepseekai2025deepseekv3technicalreport} demonstrate the feasibility of a single general-purpose model, trained on extensive text corpora, for diverse, generalizable task accomplishment \cite{reed2022generalistagent}. Capitalizing on the success of LLMs, we resort to an LLM-driven textual description approach to leverage its inherent flexibility and generalization. Leveraging fine-tuned LLMs for FL optimization starkly contrasts with those efforts, which focus on enhancing the training capabilities of LLMs in wireless FL \cite{wang2025federated,zheng2024safely}. 
However, directly applying standard LLM methodologies to this specialized domain presents several fundamental obstacles. Prominently, common approaches like training and fine-tuning require vast real-world interaction datasets, which are prohibitively time-consuming and costly to collect \cite{mower_rosllm_2024,laskar-etal-2024-systematic}. The non-availability of expert decision-making data for FL efficiency can add to the difficulty of reliably teaching an LLM to interpret wireless FL states and generate effective actions. 
Furthermore, general-purpose LLMs are inherently weak at precise mathematical reasoning and can be prone to hallucination\cite{ahn-etal-2024-large}, making them ill-suited for optimization-centric tasks, such as resource management in wireless FL. Meanwhile, the high latency associated with processing overlong token sequences for complex reasoning makes the sole adoption of LLM infeasible in time-sensitive wireless FL systems \cite{bandyopadhyay2025thinkingmachinessurveyllm}.
Therefore, due to the domain-specific expertise and computationally intensive optimization operations, extra effort is still needed to make LLMs qualified for decision-making in wireless FL.

\begin{table*}[ht]

\centering

\caption{The summary of differences with related literature on efficient FL decision-making.}

\label{table:summary}


\begin{tabularx}{\textwidth}{p{2.5cm}<{\flushleft}|p{2.0cm}<{\centering}p{2.cm}<{\centering}p{2.3cm}<{\centering}p{1.2cm}<{\centering}|X}
\hline
& Performance Guarantee & Adaptability to Environments & Artificial Design Complexity & Interaction Cost & Brief Description \\ 
\hline
\tabincell{l}{Heuristics-based\cite{273723,8761315}} & $\RIGHTcircle$ & $\CIRCLE$ & $\RIGHTcircle$ & $\CIRCLE$ & Reliance on expert hand-designed utility functions\\
\tabincell{l}{Optimization-based \\ \cite{9210812,9264742,9187798,9207871}} & $\RIGHTcircle$ & $\CIRCLE$ & $\Circle$ & $\CIRCLE$ & Mainly applicable for optimization problems with expressions \\
\tabincell{l}{RL-based \\ \cite{Zhang_Lin_Zhang_2022,10368103,10.1145/3466752.3480129,9155494,9139873,10181138,10381828,9779339}} & $\CIRCLE$ & $\Circle$ & $\RIGHTcircle$ & $\Circle$ &  Environment-specific interactions with calibrated states\\
\hline
\textbf{T-ELLM} & $\CIRCLE$ & $\RIGHTcircle$ & $\CIRCLE$ & $\CIRCLE$ & Evolution in virtual environments through GRPO with tool assistance\\
\hline
\multicolumn{6}{>{\footnotesize\itshape}r}{Notations: \rm{${\CIRCLE}$} \emph{Excellent;} \rm{${\RIGHTcircle}$} \emph{Medium;} \rm{${\Circle}$} \emph{Poor.}}
\end{tabularx}
\vspace{-0.2cm}
\end{table*}

\subsection{Contributions}
In this paper, we propose a Tool-aided Evolutionary LLM (T-ELLM) approach to realize efficient device selection and resource allocation in wireless FL. Specifically, T-ELLM adopts a natural language-based scenario prompt, where linguistic flexibility helps to improve the generalization capability\cite{pmlr-v205-ichter23a}. Afterward, on top of a mathematics-established problem decoupling, T-ELLM capably yields device selection and resource management results by answering the scenario query and invoking convex optimization tools, respectively. Furthermore, to better tackle possible environmental changes, alongside available convex optimization tools, T-ELLM builds a sample-efficient model-based virtual learning environment, which characterizes the relationship between learning effectiveness and device selection. 
Once this high-fidelity model is established, the T-ELLM agent can be trained for a vast number of episodes entirely offline, at virtually zero communication cost. In other words, since the interaction cost is entirely induced by collecting samples for the simpler virtual environment modeling, this model-based approach yields high sample efficiency by drastically reducing the communication and time overhead for practically interacting LLMs with the real world.
The main contributions of this paper are summarized as follows:
\begin{itemize}
    \item We consider a high-dimensional decision-making problem in wireless FL scenarios, and adopt a tool-based evolutionary LLM approach, which is referred to as T-ELLM. In particular, we fine-tune an LLM, which can respond to the linguistic description of wireless FL scenarios with an appropriate output of device selection.  The astonishing contextual understanding capability in LLM empowers T-ELLM to cope with environmental changes flexibly, significantly reducing manual design effort and improving adaptability.
    \item We establish the mathematical rationale to decouple the joint device selection and resource management problem, which significantly reduces the decision-making space to be learned. On this basis, augmented with contextual convex optimization-based resource management tools, T-ELLM invokes and adapts tools to perform joint optimization.
    \item We incorporate a sample-efficient model-based virtual learning environment to ground the LLM in wireless FL scenarios. In combination with the resource management tool, the model-based approach enables the LLM to learn from a simulated environment by Group Relative Policy Optimization (GRPO) \cite{shao2024deepseekmathpushinglimitsmathematical}, due to its merits in efficient trajectory collection. The continual learning in a high-fidelity simulation environment improves the decision-making capabilities of LLM, while significantly reducing communication overhead from practical interactions in the real world.

    \item We prove that the overall discrepancy between the virtual and real environment is provably bounded, yielding a limited deviation for the advantage function learned from the virtual environment from its counterpart in the real environment. Besides, the experimental results show that the proposed framework outperforms other benchmarks in terms of energy consumption, while demonstrating its adaptability to environmental changes.

\end{itemize}

\subsection{Paper Structure}

The remainder of this paper is organized as follows. We introduce the system model and problem formulation in Section \ref{sec: model}. Afterward, Section \ref{sec: decoupling} presents how to decouple the optimization problem and explains the rationale behind it. In Section \ref{too-aided LLM} and Section \ref{sec: convergence}, we propose the T-ELLM scheme and analyze its convergence properties. Afterward, we present the simulation results in Section \ref{sec: simulations}. Finally, conclusions and future works are stated in Section \ref{sec: conclusions}.

\section{System Model and Problem Formulation}
\label{sec: model}

In this section, we first present the FL model, followed by the definition of a wireless communication system model. Finally, we formulate the problem of efficient wireless FL resource management.

\begin{table}[!t]
    \centering
    \caption{{blue}List of key notations used in the paper.}
    \label{notion}
    \renewcommand{\tabularxcolumn}[1]{m{#1}}
    \begin{tabularx}{.95\linewidth}{|c|X|}
        \toprule
        \textbf{Notion}        & \textbf{Definition}  \\
        \hline
        $t$                    & Index of the rounds  \\
        \hline
        $K$                    & The total number of rounds                                      \\
        \hline
        $n$                    & Index of device      \\
        \hline
        $N$                    & Total number of devices                                         \\
        \hline
        $D_n$                  & Local dataset of device $n$                                     \\
        \hline
        $D$                    & Entire dataset       \\
        \hline
        $|D|$                  & The size of $D$      \\
        \hline
        $\bm{\omega}_n$        & Weight parameter of local model of device $n$                   \\
        \hline
        $\bm{\omega}_G$        & Weight parameter of global model                                \\
        \hline
        $\mathbb{S}_t$         & The set of indexes of the selected devices in round $t$         \\
        \hline
        $\varXi(\mathbb{S}_t)$ & Test accuracy of global model in round $t$                      \\
        \hline
        $I$                    & The number of iterations for local training                     \\
        \hline
        $C$                    & The number of CPU cycles to process one sample                  \\
        \hline
        $\zeta$                & The effective switch capacitance constant                       \\
        \hline
        $f_{t,n}$              & CPU frequency of device $n$ during the $t$-th round              \\
        \hline
        $f_{n,\text{max}}$     & The maximum CPU frequency of device $n$                         \\
        \hline
        $p_{t,n}$              & Transmission power of device $n$ during the $t$-th round         \\
        \hline
        $p_{n,\text{max}}$     & The maximum transmission power of device $n$                    \\
        \hline
        $B_{t,n}$              & Bandwidth of device $n$ during the $t$-th round                  \\
        \hline
        $B$                    & The total bandwidth of the system                               \\
        \hline
        $G_{t,n}$              & Channel coefficient of device $n$ during the $t$-th round        \\
        \hline
        $N_0$                  & The noise power spectral density                                \\
        \hline
        $s_n$                  & The size
        (in bits) of the model parameters transmitted by device $n$                              \\
        \hline
        $T^\text{cmp}_{t,n}$   & The time for local training at device n during the $t$-th round   \\
        \hline
        $E^\text{cmp}_{t,n}$   & The energy for local training at device $n$ during the $t$-th round \\
        \hline
        $T^\text{com}_{t,n}$   & The time for data uploading at device $n$ during the $t$-th round    \\
        \hline
        $E^\text{com}_{t,n}$   & The energy for data uploading at device $n$ during the $t$-th round  \\
        \hline
        $v_{t,n}$              & The transmission rate of device $n$ during the $t$-th round        \\
        \hline
        $E_t$                  & The total energy consumption during the $t$-th round             \\
        \hline
        $T_t$                  & The total time consumption during the $t$-th round               \\
        \hline
        $\mathbf{s}^m_t$       & The statistical parameters during FL training                   \\
        \hline
        $\mathbf{s}^c_t$       & State parameters related to communication and computation       \\
        \hline
        $\mathbb{R}_t$         & The allocated resources                                         \\
        \hline
        $\mathbf{a}_t$         & Joint  action of device selection and resource allocation       \\
        \hline
        $\mathbf{o}_t$         & Linguistic representations of environmental state                                          \\
        \hline
        $\tilde{\mathbf{a}}_t$ & Text-based decision action for device selection                 \\
        \bottomrule
    \end{tabularx}
\vspace{-0.5cm}
\end{table}

\subsection{System Model}
\begin{figure}[tp]
    \begin{center}
        \vspace*{-0.5em}{
            \includegraphics[width=3.5in]{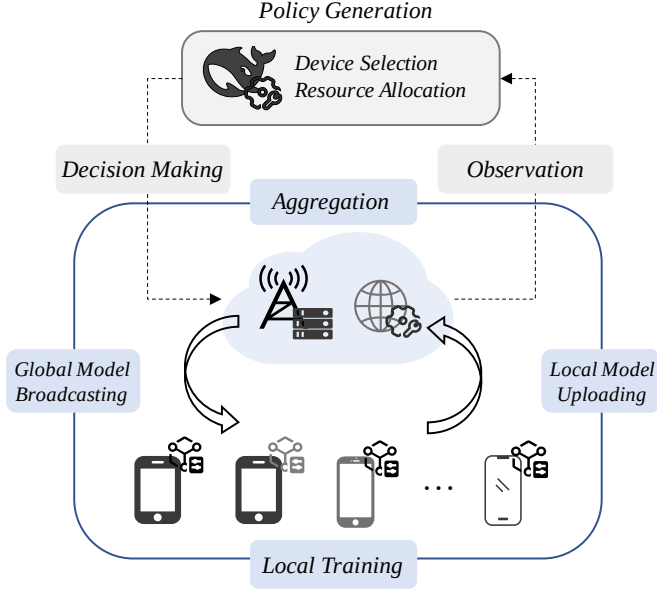}}
    \end{center}
    \vspace*{-0.5em}
    \caption{An illustration of the wireless FL environment with a device selection and resource management policy generator.}\label{fig:1.1}\vspace{-0.5em}
\end{figure}
\subsubsection{Wireless FL Model}
\label{wfm}
We consider a wireless FL system consisting of a Base Station (BS) equipped with a central server and $N$ distributed devices represented by $\mathcal{N}=\{0,1,\cdots,n,\cdots, N-1\}$. Each device $n$ has its local dataset, denoted by $D_n$, with a size $|D_n|$. For a given Neural Network (NN) training task, define the loss function as $l(\bm{\omega}, x)$, where $\bm{\omega}$ represents the parameters of the NN and $x$ is a sample in the dataset. Subsequently, the local training loss of device $n$ can be calculated as:
\begin{align}
    F_n(\bm{\omega}_n) = \frac{1}{|D_n|}\sum\nolimits_{x \in D_n}l(\bm{\omega}_n, x),
\end{align}
where $\bm{\omega}_n$ is the local parameters of device $n$. While FL aims to find global parameters $\omega_G$ such that the loss is minimized across the entire dataset $D = \bigcup_{n\in\mathcal{N}}D_n$ without sharing any datasets, the corresponding  optimization problem can be formulated as:
\begin{equation}\label{p1}
    \begin{aligned}
        \bm{\omega}^{*}_G = \arg\min_{\bm{\omega}_G}F(\bm{\omega}_G),
    \end{aligned}
\end{equation}
where $F(\bm{\omega}_G) = \sum_{n\in\mathcal{N}}\frac{|D_n|}{|D|}F_n(\bm{\omega}_G)$ is the global loss function.

As illustrated in Fig. \ref{fig:1.1}, FL algorithms typically solve \eqref{p1} through an iterative three-step process, which encompasses local training, aggregation, and synchronization, repeated over multiple rounds. Taking the example of \textit{FedAvg} \cite{mcmahan2017communication}, in the $t$-th communication round, the central server broadcasts the $(t-1)$-th global model parameters $\bm{\omega}_G^{t-1}$ to each device, where $\bm{\omega}_G^{0}$ indicates an initialized model. After getting the global model, each device $n$ uses its local dataset to update the parameters of the local model $I$ times, i.e.,
\begin{align}
    \bm{\omega}_n^t(i) = \bm{\omega}_n^t(i-1) - \eta\nabla F_n(\bm{\omega}_n^t(i-1)),
\end{align}
where $i\in[1,I]$ represents the index of local updates and $\bm{\omega}_n^t(0) = \bm{\omega}_G^{t-1}$. It is worth noting that only a portion of the devices truly participate in the training, and these selected devices in the $t$-th round are denoted as $\mathbb{S}_t \subset \mathcal{N}$. \emph{All} these selected devices transmit the trained local models to the central server, in exchange for a new, aggregated global model, i.e.,
\begin{align}\label{aggra}
    \bm{\omega}^{t}_G = \frac{1}{|D|}\sum\nolimits_{n\in \mathbb{S}_t}|D_n|\bm{\omega}_n^t,
\end{align}
where $\bm{\omega}_n^t=\bm{\omega}_n^t(I)$. Considering the impact of the subset  $\mathbb{S}_t$ on the convergence behavior of FL \cite{9252927}, the central server shall calibrate a means to determine device participation, thereby optimizing training performance.

\subsubsection{Computation and Communication Model for FL:}
\label{cacmf}
We consider the energy and time consumption associated with the computation and communication processes in the wireless FL system. 
Due to the high transmission power of the BS and sufficient downlink bandwidth for global model broadcast, the consumed time and energy at the BS are assumed to be negligible \cite{8737464,10138783}. Therefore, we only focus on the device computation and communication.
\begin{itemize}[leftmargin=*]
    \item \textit{Local Computation\cite{10181138}}:
          Let $f_{t,n}$ denote the CPU frequency (in CPU cycles per second) of device $n$ in the $t$-th round. The time required for $I$ times of local model training on device $n$ in the $t$-th round can be expressed as $T^\text{cmp}_{t,n} = \frac{C|D_{n}|I}{f_{t,n}}$,
          where $C$ (cycles per sample) represents the number of CPU cycles needed to process one sample by the backpropagation algorithm. Additionally, the energy consumption of device $n$ for local model training in the $t$-th round is given by
          $E^\text{cmp}_{t,n} =\zeta C|D_{n}|If^2_{t,n}$,
          where $\zeta$ is the effective switched capacitance that depends on the chip architecture.

    \item \textit{Model Transmission}:
          After completing local model training, the selected devices upload their local model parameters to the BS via Frequency Division Multiple Access (FDMA). The achievable transmission rate of local device $n$ in the $t$-th round is given by
          $v_{t,n}=B_{t,n}\log_2\left(1+\frac{p_{t,n}G_{t,n}}{N_0 B_{t,n}}\right)$,
          where $B_{t,n}$ and $p_{t,n}$ denotes the bandwidth and transmission power allocated to device $n$ in the $t$-th communication round respectively, $G_{t,n}$ is the channel coefficient of device $n$ in the $t$-th round under the current environmental conditions, and $N_0$ is the noise power spectral density. Let $s_n$ denote the size (in bits) of the model parameters transmitted by device $n$. Therefore, the time required for device $n$ to upload its model parameters in the $t$-th communication round is given by $T^\text{com}_{t,n} = \frac{s_n}{v_{t,n}}$.
          Subsequently, the energy consumption of device $n$ in the $t$-th round for uploading model parameters can be expressed as $E^\text{com}_{t,n} = p_{t,n}T^\text{com}_{t,n} = \frac{p_{t,k}s_n}{v_{t,n}}$.
\end{itemize}

Above all, for each communication round, the total energy consumption is modeled as:
\begin{align}
    E_t = \sum\nolimits_{n\in \mathbb{S}_t} (E^\text{com}_{t,n} + E^\text{cmp}_{t,n}).
\end{align}
Similarly, the total time consumption is given by
\begin{align}
    \label{eq:t_total}
    T_t = \max_{n\in \mathbb{S}_t}\{T^\text{com}_{t,n}+T^\text{cmp}_{t,n}\}.
\end{align}

\subsection{Problem Formulation}
We tackle the joint optimization of device selection and resource allocation in wireless FL. In particular, we aim to maximize energy efficiency while adhering to diverse computational, communication, and task-specific constraints. For example, some scenarios prioritize minimizing runtime, while others only require meeting time-related Quality-of-Service (QoS). Similarly, bandwidth allocation may be equal or flexible, contingent on the scenario.
First, we formulate an objective function that balances FL learning performance and device energy consumption, expressed as:
\begin{align}\label{e1reward}
    r(\varXi_t , E_t) = (1-\sigma)\frac{\varXi_t}{E_t}+\sigma\varXi_t,
\end{align}
where $\sigma\in[0,1]$ is a weight factor, $\varXi_t$ is the performance of the global model, i.e., test accuracy in the $t$-th round.
Subsequently, we need to optimize the objective function under resource and task requirements, and the resource constraints essentially affect resource allocation strategies. Towards an exemplified scenario, which has limited CPU operating frequency, transmission power, and bandwidth, and should satisfy some time-related QoS requirements, we optimize the Cumulative Weighted Performance-Energy Metric (CWPEM) as: 
\begin{subequations}
    \begin{align}
        \mathfrak{P}1:\quad \max_{\mathbb{S}_t, f_{t,n}, p_{t,n}, B_{t,n}} 
        & \sum\nolimits_{t=1}^{K} r(\varXi_t, E_t) \label{p1a}      \\
        \mathrm{ s.t. } \                                                  
        & T_{t} \leq T_\text{QoS},  \label{p1b}                     \\
        & f_{t,n} \leq f_{n,\text{max}} \label{p1c},                \\
        & p_{t,n} \leq p_{n,\text{max}} \label{p1d},                \\
        & \sum\nolimits_{n\in\mathbb{S}_t} B_{t,n} = B\label{p1e},
    \end{align}
\end{subequations}
where $K$ is the total round of FL, $T_\text{QoS}$ denote the QoS requirement for the FL competition time, $B$ is the total bandwidth of the system, $f_{n, \text{max}}$ and $p_{n,\text{max}}$ denote the maximum local computation capacity and maximum transmission power of device $n$, respectively. Constraint \eqref{p1b} ensures that the time consumption complies with the time-related QoS requirement. Constraints \eqref{p1c} and \eqref{p1d} specify that the allocated computational frequency and transmission power must not exceed the device's resource limitations. Finally, Constraint \eqref{p1e} guarantees that the aggregate bandwidth utilized by the selected devices adheres to the system's bandwidth constraints.

A widely applied solver to the joint optimization problem $\mathfrak{P}1$ is DRL\cite{9264742,9187798,10368103}.
Generally, the dynamic process of the wireless FL can be formally modeled as the Markov Decision Process (MDP), denoted by $\mathcal{M} = (\mathcal{S},  \mathcal{A}, P, R, \gamma) $, where $\mathcal{S}$ is the state space with $\mathbf{s}_t\in\mathcal{S}$, and $\mathbf{s}_t$ can be composed of two parts, i.e.,
\begin{align}
    \mathbf{s}_t = [\mathbf{s}^c_t, \mathbf{s}^m_t],
\end{align}
where  $\mathbf{s}^c_t$ represents parameters related to communication and computation, including  $f_{t,\text{max}}$, $p_{t,\text{max}}$, $G_{t,n}$, $ E^{\text{com}}_{t,n}$, and $ E^{\text{cmp}}_{t,n}$ in Section \ref{cacmf}, while $\mathbf{s}^m_t$ denotes the statistical parameters during training, which correspond to $|D_n|$, $F_n(\omega_n)$, and $ \varXi(\mathbb{S}_{t-1})$ in Section \ref{wfm}.
$\mathcal{A}$ is the action space with $\mathbf{a}_t\in\mathcal{A}$ encompassing device selection and corresponding resource allocation results, i.e.,
\begin{align}
    \mathbf{a}_t = [\mathbb{S}_t, \mathbb{R}_t],
\end{align}
where $\mathbb{R}_t$ represents the allocated resources, such as $f_{t,n}$, $p_{t,n}$, and $B_{t,n}$, to enable the full participation of devices within the subset  $\mathbb{S}_t$, while we call such a subset as a \emph{feasible} subset\footnote{Notably, if $\mathbb{R}_t$ can only support the participation of a subset $\hat{\mathbb{S}}_t \subset \mathbb{S}_t$ of devices, we will denote the action as $[\hat{\mathbb{S}}_t, \mathbb{R}_t]$ to avoid the ambiguity while the feasible subset is denoted as $\hat{\mathbb{S}}_t$.}.
$P$ is the transition function $P(\mathbf{s}_{t+1} | \mathbf{s}_t,\mathbf{a}_{t})$. $R$ is the reward function calculated by \eqref{e1reward} and $\gamma$ is the discount factor.
Generally, the NNs trained by DRL are only suitable for the current interactive environment, and thus, they need to be re-trained in a new environment. In addition, for different scenarios with distinctive resource constraints and task requirements, the NNs might be redesigned from scratch. For example, for two scenarios with different managed resources, the action space must be redesigned. Unfortunately, for a complicated problem that involves multiple interdependent decision-making tasks, the design and training of NNs becomes increasingly complex. Besides, since the number of neurons grows exponentially with the number of decision variables, such as $f_n$, $p_n$, and $B_n$ per device, it commonly encounters scalability issues. Consequently, an excessively large decision-making action space adversely affects the efficacy of DRL.
Therefore, we propose to introduce LLM to address these challenges.


\section{Decoupling of the Optimization Problem}
\label{sec: decoupling}
In this section, to make LLM compatible with the decision-making environment of wireless FL, we decouple the optimization problem. In particular, we decompose the joint optimization into two distinct sub-problems.
Beforehand, we show that under a feasible device selection subset $\mathbb{S}_t$, there exists independence between the global model performance and the resource allocation.

\begin{lemma}\label{srind}
    Given the feasible device selection action $\mathbb{S}_t$, the performance of global model $\varXi$ is independent of the valid resource allocation action $\mathbb{R}_t$, where $\mathbb{R}_t$ enable the full participation of devices within the subset $\mathbb{S}_t$, i.e.,
    \begin{align}
        P(\varXi_t, \mathbb{R}_t|\mathbb{S}_t) = P(\varXi_t|\mathbb{S}_t)P(\mathbb{R}_t|\mathbb{S}_t).
    \end{align}
\end{lemma}

\begin{proof}
    By definition, all devices in $\mathbb{S}_t$ are involved in the aggregation of FL, which implies no resource constraints in \eqref{p1a} are violated. In this case, the parameters of the global model are completely determined by \eqref{aggra}. Mathematically,
    \begin{align}
        P(\varXi_t|\mathbb{R}_t, \mathbb{S}_t) = P(\varXi_t|\mathbb{S}_t).
    \end{align}
    Therefore, we have
    \begin{equation}
        \begin{aligned} {P(\varXi_t,\mathbb{R}_t| \mathbb{S}_t)} & = \frac{P(\varXi_t,\mathbb{R}_t, \mathbb{S}_t)}{P(\mathbb{S}_t)}               \\
        & =\frac{P(\mathbb{R}_t, \mathbb{S}_t)P(\varXi_t|\mathbb{S}_t)}{P(\mathbb{S}_t)} \\
        & =P(\varXi_t|\mathbb{S}_t)P(\mathbb{R}_t| \mathbb{S}_t).
        \end{aligned}
    \end{equation}
    We have the lemma.
\end{proof}

\begin{corollary}
    \label{cor:independenceXiResource}
    The performance of the global model $\varXi$ is solely determined by the selected subset of participating devices $\mathbb{S}_t$, i.e.,
    \begin{align}
        \varXi_t = \varXi(\mathbb{S}_t).
    \end{align}
\end{corollary}



Afterward, we introduce some fundamental assumptions for the problem decoupling.
\begin{assumption}\label{assum1}
    For a general resource management problem with a feasible device subset $\mathbb{S}_t$, there always exists a function $g(\cdot)$ that can find the optimal solution of the joint optimization problem $\mathfrak{P}1$.
\end{assumption}
\begin{assumption}\label{assum2}
    For the joint optimization problem $\mathfrak{P}1$, given fixed and unchanging device resources (i.e., $f_{t,n} = f_{n}$, $p_{t,n} = p_{n}$ and $B_{t,n} = B_n$), an optimal feasible subset $\mathbb{S}_t^*$ can always be found.
\end{assumption}

Based on these assumptions, we have the following theorem.
\begin{theorem} \label{theo_dual}
    Under Assumption \ref{assum1} and  Assumption \ref{assum2}, the joint optimization problem $\mathfrak{P}1$ can be equivalently decomposed into an energy-efficiency oriented resource management subproblem $\mathfrak{P}2$ and a device selection subproblem $\mathfrak{P}3$, where $\mathfrak{P}2$ and $\mathfrak{P}3$ are respectively defined as:
    \begin{align}
        \mathfrak{P}2:\quad \min_{f_{t,n}, p_{t,n}, b_{t,n}} & \sum\nolimits_{t=1}^{K} E_t \nonumber \\
        \mathrm{ s.t. } \ & \mathbb{S}_t \subset \mathcal{N}, \label{p2a} \\
        & \rm{constraints} \ \eqref{p1b}, \eqref{p1c}, \eqref{p1d}, \eqref{p1e},\nonumber
    \end{align}
    and
    \begin{equation}
        \begin{aligned}
            \mathfrak{P}3:\quad & \max_{\mathbb{S}_t} \sum\nolimits_{t=1}^{K} r(\varXi (\mathbb{S}_t), E_t^*(\mathbb{S}_t)) \\
            \mathrm{ s.t. } \   & T_{t,n} \leq T_\text{QoS}, \\
                                & f_{t,n}, p_{t,n}, B_{t,n} =  g(\mathbb{S}_t).
        \end{aligned}
    \end{equation}
\end{theorem}
\begin{proof}
    We first discuss the resource management issue in the problem $\mathfrak{P}1$. Assumption \ref{assum1} implies that for a given feasible device subset $\mathbb{S}_t$, through $g(\cdot)$, the optimal resources can always be allocated to the device to minimize energy consumption, while meeting the required QoS and resource constraints. In other words, given any device subset $\mathbb{S}_t$, the minimum energy consumption can be calculated as $E_t^*(\mathbb{S}_t) = E_t(\mathbb{S}_t,g(\mathbb{S}_t))$ as in the problem $\mathfrak{P}2$.

    Next, recalling the optimization objective $r(\varXi (\mathbb{S}_t), E_t)$ in the problem $\mathfrak{P}1$, it monotonically decreases with respect to $E_t$ and increases with respect to $\varXi (\mathbb{S}_t)$,  since $\varXi$ is independent of resource allocation by Corollary \ref{cor:independenceXiResource}. Based on Assumption \ref{assum2}, the optimal feasible subset $\mathbb{S}_t^*$ can always be found for the sub-problem $\mathfrak{P}3$. Afterward, for any given device subset $\mathbb{S}_t^*$, it follows that
    \begin{equation}
        \begin{aligned}
              & \max_{\mathbb{S}_t, f_{t,n}, p_{t,n}, b_{t,n}}\sum\nolimits_{t=1}^{K} r(\varXi (\mathbb{S}_t), E_t) \\   =& \max_{f_{t,n}, p_{t,n}, b_{t,n}}\sum\nolimits_{t=1}^{K} r(\varXi (\mathbb{S}_t^*), E_t) \\
            = & \sum\nolimits_{t=1}^{K} r(\varXi (\mathbb{S}_t^*), E_t^*(\mathbb{S}_t^*)).
        \end{aligned}
    \end{equation}
    Therefore, $\mathfrak{P}1$ is equivalent to solving $\mathfrak{P}2$ and $\mathfrak{P}3$ separately, and the optimal solution is given by $(\mathbb{S}_t^*, g(\mathbb{S}_t^*))$.
\end{proof}

\noindent
\textit{Remark}: Theorem \ref{theo_dual} provides the formal basis for decoupling the joint optimization problem. This allows us to deal with the two sub-problems separately and reduces computational complexity by searching over a smaller-dimensional solution space. Accordingly, we can introduce LLM to solve $\mathfrak{P}3$ conveniently, while resorting to a tool-aided solution of $\mathfrak{P}2$. Notably, both sub-problems still involve a calibrated joint design for decision making: For the subproblem $\mathfrak{P}2$, the function $g(\cdot)$ satisfying Assumption \ref{assum1} can be practically guaranteed by optimization tools. For $\mathfrak{P}2$, the policy itself is required to make optimal device selection decisions under the condition of the resource allocation mechanism $g(\cdot)$.

\section{Tool-Aided Evolutionary LLM Scheme}
\label{too-aided LLM}
In this section, we first introduce the tool-aided LLM towards generating an effective device selection and resource allocation policy. Afterward, on top of
convex optimization-based resource management tools, we introduce a model-based, tool-assisted training framework. Finally, we discuss how to train the LLM by GRPO, thus improving its decision-making capabilities.
\begin{figure}[tp]
    \begin{center}
        \vspace*{-0.5em}{
            \includegraphics[width=3.5in]{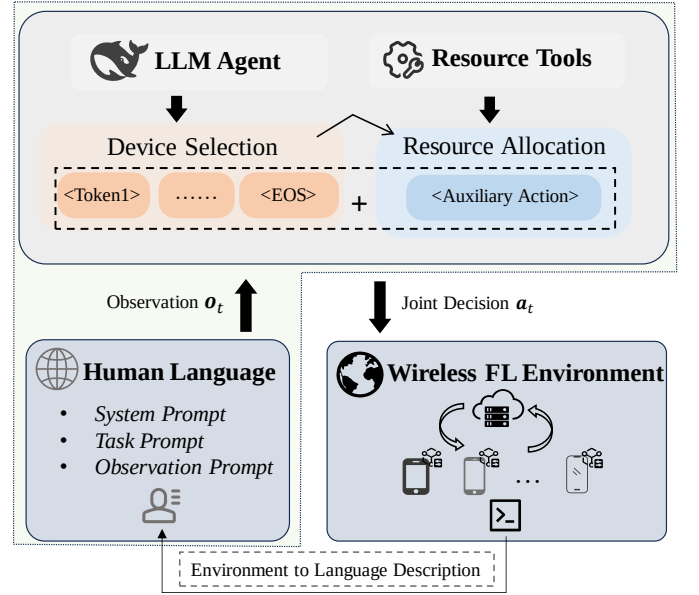}}
    \end{center}
    \vspace*{-0.5em}
    \caption{An illustration of the inference process by the tool-aided  LLM.}\label{fram}\vspace{-0.5em}
\end{figure}

\subsection{Tool-Aided LLM for Efficient Policy Generation}
The decision-making pipeline of the proposed tool-aided LLM is shown in Fig. \ref{fram}, which is motivated by the mathematical findings in Section \ref{sec: decoupling} and can be divided into two parts: a generalizable LLM to provide device section results for $\mathfrak{P}3$ and an optimization tool to yield scenario-specific contextual resource allocation outcomes for $\mathfrak{P}2$.
To generate a resource-efficient policy, at each communication round $t$, the current observation of the FL environment will be captured and described directly in language.
Correspondingly, the state space $\mathcal{S}$ in the MDP  $\mathcal{M}$ is converted to text-based observation $\mathcal{O}$ with $\mathbf{o}_t\in \mathcal{O}$, and
\begin{align}\label{observa}
    \mathbf{o}_t =[\mathbf{s}^c_t, \mathbf{s}^m_t, \mathbf{s}^{\text{text}}_t],
\end{align}
where $\mathbf{s}^{\text{text}}_t$ represents the added natural language description.
This linguistic representation $\mathbf{o}_t$, which will be elaborated later, serves as the input to the LLM. Consequently, it efficiently adapts to diverse state spaces across various scenarios without requiring additional tuning of NN architectures.  Let $\mathcal{V}$ represent the LLM's entire token space. For a given observation $\mathbf{o}_t$, the LLM first computes a vector of logits over all tokens in $\mathcal{V}$, i.e., $L_v(\mathbf{o_t})$. To ensure that the generated action corresponds to a valid device, we define a subset of this token space $\mathcal{I}\subset \mathcal{V}$, as the set of valid action tokens, where each token represents a unique device index $n\in \mathcal{N}$. Therefore, the logits over all valid tokens can be expressed as:
\begin{equation}
  \begin{aligned} 
L'_{v}(\mathbf{o}_t) = \begin{cases} L_{v}(\mathbf{o}_t) & \text{if } v \in \mathcal{I} \\ -\infty & \text{if } v \notin \mathcal{I} \end{cases} .
\end{aligned}  
\end{equation}
A softmax function is then applied to this masked logit vector to produce the final probability distribution over the vocabulary, that is, $\pi_\theta(\cdot|\mathbf{o}_t) := \pi_\theta(v|\mathbf{o}_t) = \frac{\exp(L'_{v}(\mathbf{o}_t))}{\sum_{u \in \mathbb{V}} \exp(L'_{u}(\mathbf{o}_t))}$.
Then, the LLM generates a text response $\tilde{\mathbf{a}}_{t}$ based on the current language description according to its strategy $\pi_{\theta}$, i.e.,
\begin{align}
    \tilde{\mathbf{a}}_{t} \sim \pi_\theta(\cdot|\mathbf{o}_t),
\end{align}
which is interpreted as the decision action for device selection $\mathbb{S}_t$.
Upon selecting the subset $\mathbb{S}_t$ of devices, the LLM invokes an external resource \emph{tool} to derive the resource allocation action $\mathbb{R}_t$ for the selected devices. Finally, the joint decisions $\mathbf{a}_t = [\mathbb{S}_t, \mathbb{R}_t]$ are applied to the wireless FL environment. Thus, the next observation can be obtained by the environment, i.e.,
\begin{align}
    P(\mathbf{o}_{t+1}|\mathbf{o}_t,\mathbf{a}_t).
\end{align}
Afterward, the feedback provided to the LLM-based policy generator triggers the next round of efficiency optimization.
\subsubsection{The LLM for $\mathfrak{P}3$}

As shown in Fig. \ref{llm_example}, the LLM takes the linguistic description $\mathbf{o}_t$ of the environmental observations as input and generates the device selection decisions $\mathbb{S}_t$, corresponding to the indices of devices. Prominently, the prompts, which include system prompt, task prompt, and observation prompt, are specially tailored for the wireless FL problem $\mathfrak{P}1$.
Notably, system and task prompts configure the LLM as an agent for FL. These prompts guide the LLM in analyzing problems based on the provided task description, thereby increasing the likelihood of generating task-related token outputs.
Besides, the observation prompt integrates all the state parameters $\mathbf{s}^c_t$ and $\mathbf{s}^m_t$ required for the current task, furnishing the LLM with a comprehensive description of the current environment for decision-making.
Additionally, to ensure compliance with operational constraints, the prompt also specifies the desired output format of $\mathbb{S}_t$ as a token-based response $\tilde{\mathbf{a}}_{t}$. On this basis, we fine-tune the LLM to improve the performance.
\begin{figure*}[tp]
    \centering
    \includegraphics[width=1\linewidth]{figs//example_short.pdf}
    \caption{An example of input and output for device selection decision-making
        by LLM.}\label{llm_example}\vspace{-2.0em}
\end{figure*}
\subsubsection{The Tool for $\mathfrak{P}2$}
For the tool-aided LLM, the tool shall have the capability to compute optimal resource allocation solutions to $\mathfrak{P}2$ under a unique scenario, so that LLM can be compatible with the specific tool to make efficient joint decisions in response to this specific scenario. Numerous qualified algorithms \cite{9264742,9187798} can be leveraged. In this work, due to its implementation simplicity, we adopt the alternative direction algorithm ALTD proposed by \cite{9187798}. Notably, for $\mathfrak{P}2$, based on device selection $\mathbb{S}_t$, ALTD optimizes allocated resources $\mathbb{R}_t$ including the CPU frequency $f_{t,n}$, transmission power $p_{t,n}$ and bandwidth allocation $B_{t,n}$ of networking devices under the constraints of average bandwidth allocation and QoS requirements, thus allowing the participating devices to perform FL with the lowest energy consumption.
\subsection{Model-Based Virtual Environment}
\label{ModelVE}
Due to the significant communication overhead associated with online interactions in wireless FL, this section proposes a virtual environment based on an offline model-based approach and tool-assisted computations. Benefiting from the following lemma, the learning of the virtual environment can avoid the extensive use of huge end-to-end expert data.
\begin{lemma}
    \label{syssp}
    The state transition of the wireless FL environment can be decomposed into two parts:
    \begin{align}
        \label{eq: syssp}
        P(\mathbf{s}_{t+1}|\mathbf{s}_{t},\mathbf{a}_{t}) = P(\mathbf{s}^{c}_{t+1}|\mathbf{s}^{c}_{t},(\mathbb{S}_{t},\mathbb{R}_{t}))P(\mathbf{s}^{m}_{t+1}|\mathbf{s}^{m}_{t},\mathbb{S}_{t}).
    \end{align}
\end{lemma}
\begin{proof}
    The state transition of the wireless RL environment can be represented as:
    \begin{equation}
        \begin{aligned}
            P(\mathbf{s}_{t+1}|\mathbf{s}_{t},\mathbf{a}_{t}) 
            & = P(\mathbf{s}^{m}_{t+1},\mathbf{s}^{c}_{t+1}|\mathbf{s}_{t},\mathbf{a}_{t}) \\
            & = P(\mathbf{s}^{m}_{t+1},\mathbf{s}^{c}_{t+1}|(\mathbf{s}^{m}_{t},\mathbf{s}^{c}_{t}),(\mathbb{S}_{t},\mathbb{R}_{t})).
        \end{aligned}
    \end{equation}
    By definition, $\mathbf{s}^{c}$ and $\mathbf{s}^{m}$ are conditional independent. Therefore, we can have
    \begin{equation}
        \begin{aligned}
              & P(\mathbf{s}^{m}_{t+1},\mathbf{s}^{c}_{t+1}|(\mathbf{s}^{m}_{t},\mathbf{s}^{c}_{t}),(\mathbb{S}_{t},\mathbb{R}_{t}))\\
            = & P(\mathbf{s}^{m}_{t+1}|(\mathbf{s}^{m}_{t},\mathbf{s}^{c}_{t}),(\mathbb{S}_{t},\mathbb{R}_{t}))P(\mathbf{s}^{c}_{t+1}|(\mathbf{s}^{m}_{t},\mathbf{s}^{c}_{t}),(\mathbb{S}_{t},\mathbb{R}_{t})).
        \end{aligned}
    \end{equation}
    Similarly, according to the independence of $\mathbf{s}^{m}_{t+1}$ and $\mathbf{s}^{c}_{t}$, as well as $\mathbf{s}^{c}_{t+1}$ and $\mathbf{s}^{m}_{t}$, we have 
    \begin{align}  P(\mathbf{s}_{t+1}|\mathbf{s}_{t},\mathbf{a}_{t}) = P(\mathbf{s}^{m}_{t+1}|\mathbf{s}^{m}_{t},(\mathbb{S}_{t},\mathbb{R}_{t}))P(\mathbf{s}^{c}_{t+1}|\mathbf{s}^{c}_{t},(\mathbb{S}_{t},\mathbb{R}_{t})).
    \end{align}
    Recalling that by definition, $\mathbf{s}^{m}$ is $\{ |D_n|, F_n(\omega_n), \varXi(\mathbb{S}_{t-1})\}$-related. Then, by Lemma \ref{srind}, we have the conclusion.
\end{proof}

\begin{figure*}[t]
    \centering
    \includegraphics[width=1\linewidth]{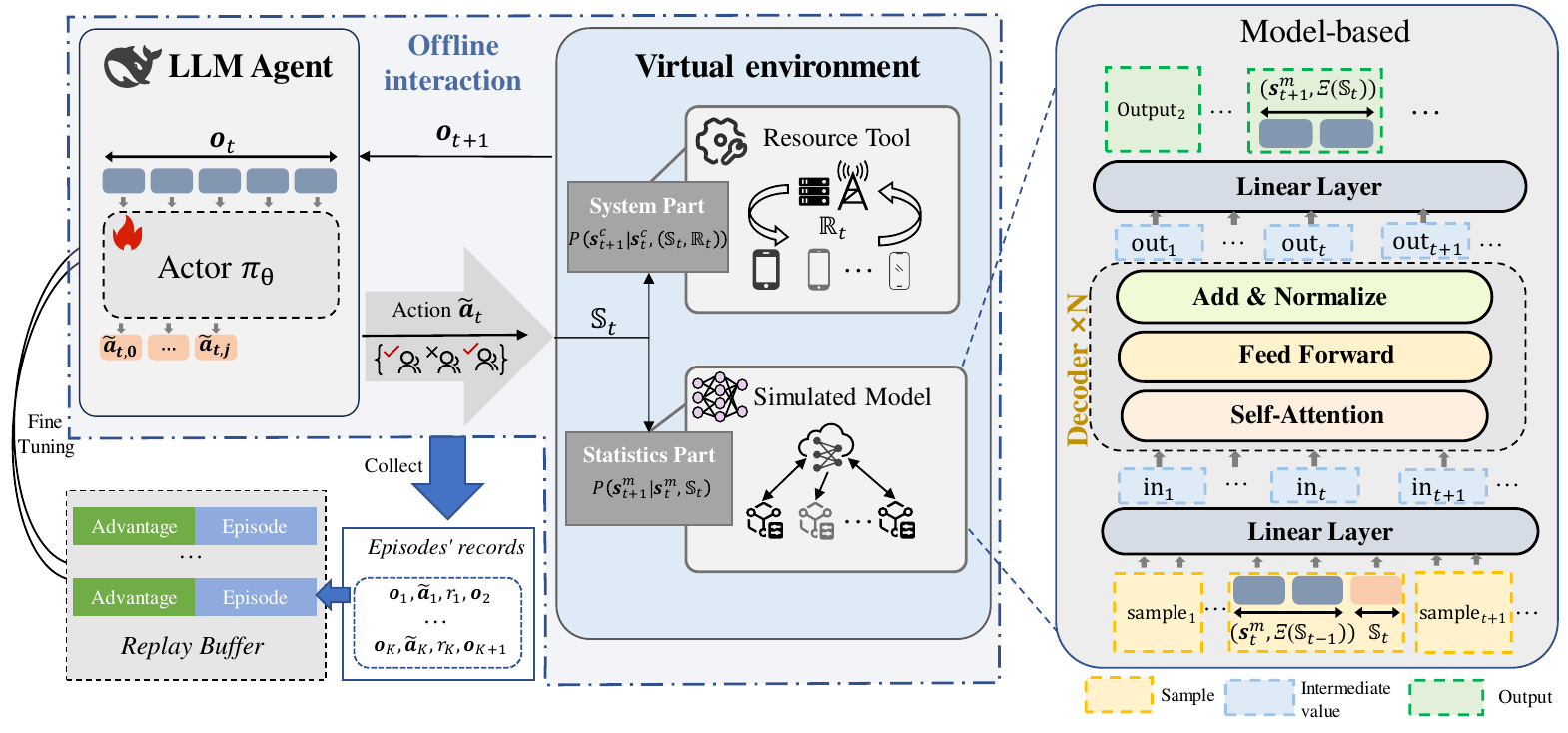}
    \caption{Illustration of the evolutionary virtual environment and GRPO-based training in T-ELLM.}\label{model_view}
    \vspace{-1.0em}
\end{figure*}

Lemma \ref{syssp} implies that the wireless FL environment can be simulated by a \emph{system part} and an FL \emph{statistics part}, which is illustrated in Fig. \ref{model_view}.
For the system part, it only focuses on the communication and computation status of devices according to the given resources, i.e., $P(\mathbf{s}^{c}_{t+1}|\mathbf{s}^{c}_{t},(\mathbb{S}_{t},\mathbb{R}_{t}))$. Therefore, the observation and consumption can be recorded and simulated by the resource tool. Specifically, based on $\mathbb{R}_t$
and $\mathbb{S}_t$, we can leverage the tool to calculate the consumed time $T_t$ and energy $E_t$, and update the state $\mathbf{s}^c_{t+1}$.

However, for characterizing the statistics part $P(\mathbf{s}^{m}_{t+1}|\mathbf{s}^{m}_{t},\mathbb{S}_{t})$, it remains challenging when dynamically selected devices participate in training and aggregation. To address this, we employ a model to approximate the convergence behavior and performance of the global model during FL training. In other words, simulating the statistics part $P(\mathbf{s}^{m}_{t+1}|\mathbf{s}^{m}_{t},\mathbb{S}_{t})$ means to predict $\mathbf{s}^{m}_{t+1}$ including $\varXi(\mathbb{S}_{t})$ based on the current observation $\mathbf{s}^{m}_{t}$, device selection action $\mathbb{S}_t$ and the historical accuracy $\varXi(\mathbb{S}_{t-1})$. Generally, $\mathbf{s}^{m}_{t}$ can be classified as the state information pertinent to the selected devices $\mathbf{s}^{m}_t(s)$ and that of all devices $\mathbf{s}^{m}_t(l)$. In particular, $\mathbf{s}^{m}_t(s)$ includes post-training local loss $F_n(\bm{\omega}_n)$, the inner-product $\langle \nabla F_n(\bm{\omega}_n^t), \nabla F_G(\bm{\omega}_G^t)\rangle$, and the percentage of the same sign $e(\bm{\omega}_n^t, \bm{\omega}_G^t)$ shared between local and global models (i.e., $\bm{\omega}_n^t$ and $\bm{\omega}_G^t)$) for all selected devices $n \in \mathbb{S}_{t}$. Meanwhile, $\mathbf{s}^{m}_t(l)$ consists of the local dataset size $|D_n|$ (which remains constant during training) and locally computed loss $l_n$ of other non-selected devices. We denote the tuple $((\mathbf{s}^{m}_t, \varXi(\mathbb{S}_{t-1}), \mathbb{S}_{t}),(\mathbf{s}^{m}_{t+1}, \varXi(\mathbb{S}_{t})))$ as one sample of the environment model, while several continuous samples constitute a trajectory. The NN architecture of the simulated model is shown in the right part of Fig. \ref{model_view}, while
 the details of the training process are shown in Algorithm \ref{alg1}.
During training, we optimize the model with a Mean Squared Error (MSE) loss.

Generally, the reward $r_t$ can be computed as in \eqref{e1reward},  except when the resource tool finds no solution exists for $\mathfrak{P}$2, the action $\mathbb{S}_t$ will be given a penalty, i.e., $r_t =0$. Then, alongside the reward $r_t$, the virtual environment can respond to $(\mathbf{s}^{m}_t, \varXi(\mathbb{S}_{t-1}), \mathbb{S}_{t})$, and accurately yield the next state $(\mathbf{s}^c_{t+1}, \mathbf{s}^m_{t+1})$.

\subsection{GRPO-Based LLM Training}
The aforementioned virtual environment lays the very foundation for updating the LLM agent in an RL manner. Specifically, we employ GRPO \cite{shao2024deepseekmathpushinglimitsmathematical} to enhance the inference capability of LLM. As a variant of Proximal Policy Optimization (PPO) \cite{schulman_proximal_2017}, GRPO also adopts twin policy models, where an old policy $\pi_{\theta_\text{old}}$ is used for sampling actions $\tilde{\mathbf{a}}_{t}$ and accumulating transitional records $(\mathbf{o}_t, \tilde{\mathbf{a}}_{t}, r_t, \mathbf{o}_{t+1})$ in a replay buffer. Here, $r_t$ is the reward in the $t$-th round calculated as in Section \ref{ModelVE}.
All samples are described in language, a specific example as shown in Fig. \ref{llm_example}. For a batch of $K$-length episodes' records, where each episode $i$ is denoted as $\{r_1,\cdots,r_t,\cdots,r_K\}_i$, we normalize these rewards with the $t$-wise average and standard deviation across episodes within the batch, i.e., $\{\tilde{r_t}\}_i = \frac{\{r_t\}_i-\text{mean}}{\text{std}}$. Then the $t$-th advantage in $i$-th episode can be calculated as:
\begin{align}\label{advgrpo}
    \{A_t\}_i = \sum\nolimits_{j\geq t} \{\tilde{r_j}\}_i.
\end{align}

Based on the $\pi_{\theta_\text{old}}$-induced records in the relay buffer, it becomes ready to update the current policy model $\pi_{\theta}$. Notably, due to the autoregressive characteristics of language tasks, GRPO \cite{shao2024deepseekmathpushinglimitsmathematical} can regressively update LLMs by taking a single token as an action. While in a wireless FL environment, an effective action is the entire response rather than a token. Therefore, wireless FL is completely different from natural language processing tasks. 
Let $\mathbf{a}_{t,j}$ represent the action taken in round $t$ and $j$-th token, thus $\mathbf{o}_{t,<j} = [\mathbf{o}_{t,<j-1}, \tilde{\mathbf{a}}_{t,j}]$, and  $\mathbf{o}_{t,<0} = \mathbf{o}_{t}$.
In this case, the conditional probability of each action $\tilde{\mathbf{a}}_{t}$ can be calculated as:
\begin{align}
    \pi_{\theta}(\tilde{\mathbf{a}}_{t}|\mathbf{o}_{t}) =\prod_{j}  \pi_{\theta}(\tilde{\mathbf{a}}_{t,j}|\mathbf{o}_{t,<j-1}).
\end{align}
Next, we adopt GRPO to optimize the LLM by maximizing
\begin{equation}\label{ppoobj}
    \begin{split}
        J_\text{GRPO}(\theta) &= \mathbb{E} \bigg[ \min\bigg(
            \frac{\pi_{\theta}(\tilde{\mathbf{a}}_{t}|\mathbf{o}_{t})}{\pi_{\theta_\text{old}}(\tilde{\mathbf{a}}_{t}|\mathbf{o}_{t})} {A}_t, \\
            & \text{clip}\left( \frac{\pi_{\theta}(\tilde{\mathbf{a}}_{t}|\mathbf{o}_{t})}{\pi_{\theta_\text{old}}(\tilde{\mathbf{a}}_{t}|\mathbf{o}_{t})}, 1-\epsilon,1+\epsilon \right) {A}_t
            \bigg) \bigg], 
    \end{split}
\end{equation}
where $\epsilon$ is a clipping-related hyperparameter for stabilizing training.

Finally, the implementation process of the T-ELLM scheme is illustrated in \textbf{Algorithm \ref{alg1}}.
\begin{algorithm}[!t]
    \caption{GRPO-based T-ELLM scheme in the model-based virtual environment.}\label{alg1}
    {\bf Training for model-based virtual environment}\\
    \hspace*{0.02in} Collect $T_{m}$ trajectories $[((\mathbf{s}^{m}_t, \varXi(\mathbb{S}_{t-1}), \mathbb{S}_{t}),(\mathbf{s}^{m}_{t+1}, \varXi(\mathbb{S}_{t})))]$ as the training dataset.
    \algblock{ParFor}{EndParFor}
    \algnewcommand\algorithmicparfor{\textbf{parfor}}
    \algnewcommand\algorithmicpardo{\textbf{do}}
    \algnewcommand\algorithmicendparfor{\textbf{end\ parfor}}
    \algrenewtext{ParFor}[1]{\algorithmicparfor\ #1\ \algorithmicpardo}
    \algrenewtext{EndParFor}{\algorithmicendparfor}
    \begin{algorithmic}[1]
        \State Add two linear layers (i.e., \texttt{linear1} and \texttt{linear2}) for embedding to GPT-2 model and initialize all parameters $\phi$.
        \For {each epoch}
        \For {each batch}
        \ParFor {all samples in the same trajectory}
        \State$\text{in}_t = \texttt{linear1}(\mathbf{s}^{m}_t, \varXi(\mathbb{S}_{t-1}), \mathbb{S}_{t})$.
        \State $\text{out}_t = \texttt{GPT}(\text{in}_1,\cdots,\text{in}_{t-1})$. 
        \State $({blue}\hat{\mathbf{s}}^m_{t+1},\hat{\varXi}(\mathbb{S}_{t})) = \texttt{linear2}(\text{out}_t)$.
        \State $\text{loss}_t = \texttt{MSE}((\hat{\mathbf{s}}^m_{t+1},\hat{\varXi}(\mathbb{S}_{t})), ({\mathbf{s}}^m_{t+1},{\varXi}(\mathbb{S}_{t})))$.
        \EndParFor
        \State $\text{loss}_\text{traj} = \texttt{mean}(\text{loss}_t)$.
        \EndFor
        \State \textbf{end for}
        \State $\text{loss}_\text{batch} = \texttt{mean}(\text{loss}_\text{traj})$.
        \State $\text{loss}_\text{batch}$ backward.
        \State update $\phi$ and two linear layers.
        \EndFor
        \State \textbf{end for}
    \end{algorithmic}

    {\bf Fine-tuning for tool-aided evolutionary LLM}\\
    \hspace*{0.02in} Initialize all the parameters: policy model $\pi_{\theta}$,  hyperparameters $\epsilon$, $\lambda$, the trained simulated model.
    \begin{algorithmic}[1]
        \State Initialize  $\pi_{\theta_\text{old}} = \pi_{\theta}$.
        \For {each iteration}
        \State $\pi_{\theta_\text{old}} = \pi_{\theta}$.
        \State Episode set $\text{Epis} = \varnothing$.
        \For {$i=1,\cdots, \text{batch}$}
        \State $\text{Epi}_i = \varnothing$
        \For {communication round $t=1,\cdots, K$}
        \State Obtain the observation $\mathbf{o}_t$.
        \State Sample $\mathbf{a}_t \sim \pi_{\theta_\text{old}}(\cdot|\mathbf{o}_t)$.
        \State Execute $\mathbf{a}_t$, receive $r_t$.
        \State Get $\mathbf{o}_{t+1}$ via resource tool and simulated model.
        \State $\text{record}_t = (\mathbf{o}_t, \tilde{\mathbf{a}}_{t}, r_t, \mathbf{o}_{t+1})$
        \State  Store $\text{Epi}_i \gets \text{Epi}_i \cup \text{record}_t$.
        \EndFor
        \State \textbf{end for}
        \State $\text{Epis} \leftarrow \text{Epis}\bigcup \text{Epi}_i$.
        \EndFor
        \State \textbf{end for}
        \State Calculate all advantages $A_t$ based on \eqref{advgrpo}.
        \For {each update step}
        \State Update  $\pi_{\theta}$ by maximing \eqref{ppoobj} with  $\text{Epis}$.
        \EndFor
        \State \textbf{end for}
        \EndFor
        \State \textbf{end for}
    \end{algorithmic}
\end{algorithm}

\section{Theoretical Accuracy of the Policy Learned from the Virtual Environment}
\label{sec: convergence}
This section theoretically validates the proposed model-based virtual environment by establishing a generalized simulation lemma. Specifically, we demonstrate that the overall discrepancy between the simulated environment and the real environment is provably bounded, i.e., the convergence properties of FL can be faithfully replicated under the model-based environment. Therefore, we further show that the advantage function, i.e., cumulative normalized reward under the learned virtual environment in \eqref{advgrpo}, remains within a bounded deviation from the real environment, ensuring similar convergence behavior of LLMs in both scenarios.


The MDP of the entire wireless FL system is expressed as $\mathcal{M} = (\mathcal{O},  \mathcal{A}, P, \mathcal{R}, \gamma)$. Specially, the state space is text-based observation $\mathbf{o}_t =[\mathbf{s}^c_t, \mathbf{s}^m_t, \mathbf{s}^{\text{text}}_t]$. 
And the added natural language description $\mathbf{s}^{\text{text}}_t$ as the prompt for LLM is independent of the wireless FL.
Similarly, let $\hat{\mathcal{M}}$ denote the MDP representing the complete model-based virtual environment, and $\hat{P}$ its corresponding transition function.
To formally characterize the difference between the real environment $\mathcal{M}$ and the model-based approximation $\hat{\mathcal{M}}$, we first establish the following assumptions and define the return-based advantage function used in our analysis.

\begin{assumption}
    \label{as: bounded_r_error}
    Let $\mathcal{M}$ and $\hat{\mathcal{M}}$ be two MDPs defined over the same state and action spaces. Suppose that the normalized reward functions satisfy the following uniform deviation bound for all state-action pairs $(\mathbf{o}_t, \mathbf{a}_t)$, the reward functions satisfy
    \begin{align}
        |\tilde{r}_{\hat{\mathcal{M}}}(\mathbf{o}_t,\mathbf{a}_t) - \tilde{r}_{\mathcal{M}}(\mathbf{o}_t,\mathbf{a}_t)|<\tilde{R}_{\max},
    \end{align}
    where $\tilde{R}_{\max}$ denotes a constant.
\end{assumption}

\begin{definition}
    For any policy $\pi$ and state $\mathbf{o}_t$, consistent with \eqref{advgrpo}, define the return-based advantage function as:
    \begin{align}
        A^\pi_P(\mathbf{o},\mathbf{a}) := \mathbb{E}_{P, \pi}\left[\sum\nolimits_{j=t}^{K-1} \tilde{r}\left(\mathbf{o}_j, \mathbf{a}_j\right)\right],
    \end{align}
    where $(\mathbf{o}_t, \mathbf{a}_t) = (\mathbf{o},\mathbf{a})$, $\tilde{r}(\mathbf{o},\mathbf{a})$ denotes a normalized reward function satisfying $\left|\tilde{r}(\mathbf{o},\mathbf{a})\right| \le \tilde{R}_{\max}$.
\end{definition}

Our approach separately characterizes the system and statistics parts instead of using a real wireless FL environment. The system part uses resource tools to simulate communication and computing costs, where fixed resource (e.g., bandwidth and transmission power) settings often lead to deterministic state transitions $P(\mathbf{s}^{c}_{t+1}|\mathbf{s}^{c}_{t},(\mathbb{S}_{t},\mathbb{R}_{t}))$, due to the explicit mathematical expressions in Section \ref{cacmf}. On the other hand, the MDP corresponding to the statistical component is denoted as $\mathcal{M}_d = (\mathcal{S}^m, \mathcal{A}^m, P^m, \mathcal{R}^m, \gamma)$, where, as described in Section \ref{too-aided LLM}, the state is defined as $\mathbf{s}^{m}_{t} = \{\mathbf{s}^{m}_{t}(s),\mathbf{s}^{m}_{t}(l)\} \in \mathcal{S}^m$. The decision action corresponds to selecting a subset $\mathbb{S}_t$ of devices. The transition probability is given by $P^m=P(\mathbf{s}^{m}_{t+1}|\mathbf{s}^{m}_{t},\mathbb{S}_{t})$, and the reward is defined as the accuracy of the global model, denoted by $\varXi(\mathbb{S}_t)$. To approximate this statistical MDP, we employ a trained model $\hat{\mathcal{M}}_d$, whose transition dynamics are captured by the learned function $\hat{P}^m$.
\begin{assumption}[Bounded Transition Discrepancy in TV Distance]
    \label{as: bounded_p_error}
    Let $P^m$ and $\hat{P}^m$ denote the transition functions of $\mathcal{M}_d$ and $\hat{\mathcal{M}}_d$, respectively. Suppose the transition model discrepancy is bounded in Total Variation (TV) as:
    \begin{align}
        \sup_{\mathbf{s}^{m}_{t},\mathbb{S}_t} D_{\mathrm{TV}} \left( \hat{P}(\mathbf{s}^{m}_{t+1}|\mathbf{s}^{m}_{t},\mathbb{S}_t) , P(\mathbf{s}^{m}_{t+1}|\mathbf{s}^{m}_{t},\mathbb{S}_t) \right) \le \varepsilon,
    \end{align}
    where $D_{\mathrm{TV}}(p, q)$ denotes the TV \cite{devroye2018total} distance between probability distributions $p$ and $q$.
\end{assumption}

\begin{lemma}
    \label{c:bounded_p_error}
    Under Assumption \ref{as: bounded_p_error}, let $P$ and $\hat{P}$ denote the transition functions of $\mathcal{M}$ and $\hat{\mathcal{M}}$, respectively. The transition model discrepancy is bounded in total variation as
    \begin{align}
        \sup_{\mathbf{o}_{t},\mathbf{a}_t} \left\| \hat{P}(\mathbf{o}_{t+1}|\mathbf{o}_{t},\mathbf{a}_t)-P(\mathbf{o}_{t+1}|\mathbf{o}_{t},\mathbf{a}_t) \right\|_1 \le 2\varepsilon.
    \end{align}
\end{lemma}

\begin{proof}
    By definition, the transition probability function $P$ of the full environment model can be decomposed as:
    \begin{align}
        P(\mathbf{o}_{t+1}|\mathbf{o}_{t},\mathbf{a}_t)=P(\mathbf{s}^{m}_{t+1},\mathbf{s}^{c}_{t+1},\mathbf{s}^{\text{text}}_{t+1},|(\mathbf{s}^{m}_{t},\mathbf{s}^{c}_{t},\mathbf{s}^{\text{text}}_{t}),(\mathbb{S}_{t},\mathbb{R}_{t})),\nonumber
    \end{align}
    where $\mathbf{s}^{\text{text}}_t$ denotes the language inputs to the LLM, which is independent of other variables and can thus be treated as a constant.
    Accordingly, consistent with Lemma \ref{syssp}, the transition function $P$ can be simplified without loss of theoretical generality as follows:
    \begin{equation}       
    \begin{aligned}
        P(\mathbf{o}_{t+1}|\mathbf{o}_{t},\mathbf{a}_t) & =P(\mathbf{s}^{m}_{t+1},\mathbf{s}^{c}_{t+1}|(\mathbf{s}^{m}_{t},\mathbf{s}^{c}_{t})(\mathbb{S}_{t},\mathbb{R}_{t})) \\
        & = P(\mathbf{s}^{c}_{t+1}|\mathbf{s}^{c}_{t},(\mathbb{S}_{t},\mathbb{R}_{t}))P(\mathbf{s}^{m}_{t+1}|\mathbf{s}^{m}_{t},\mathbb{S}_{t}),
    \end{aligned}
    \end{equation}
    where $P(\mathbf{s}^{c}_{t+1}|\mathbf{s}^{c}_{t},(\mathbb{S}_{t},\mathbb{R}_{t}))$ is the transition function of system part, which is determined by the resource tool.
    Similarly, the transition function $\hat{P}$ in the virtual environment can be obtained as:
    \begin{align}
        \hat{P}(\mathbf{o}_{t+1}|\mathbf{o}_{t},\mathbf{a}_{t}) = P(\mathbf{s}^{c}_{t+1}|\mathbf{s}^{c}_{t},(\mathbb{S}_{t},\mathbb{R}_{t}))\hat{P}(\mathbf{s}^{m}_{t+1}|\mathbf{s}^{m}_{t},\mathbb{S}_{t}).\nonumber
    \end{align}
    Therefore, under Assumption \ref{as: bounded_p_error}, we have
    \begin{align}
        & \left\| \hat{P}(\mathbf{o}_{t+1}|\mathbf{o}_{t},\mathbf{a}_t)-P(\mathbf{o}_{t+1}|\mathbf{o}_{t},\mathbf{a}_t) \right\|_1 \nonumber
        \\=&\sum_{\mathbf{s}_{t+1}^c, \mathbf{s}_{t+1}^m}\Big|P\left(\mathbf{s}_{t+1}^c \mid \mathbf{s}_t^c, (\mathbb{S}_t, \mathbb{R}_t)\right) \cdot \hat{P}\left(\mathbf{s}_{t+1}^m \mid \mathbf{s}_t^m, \mathbb{S}_t\right) \nonumber\\
        & -P\left(\mathbf{s}_{t+1}^c \mid \mathbf{s}_t^c, (\mathbb{S}_t, \mathbb{R}_t)\right) \cdot P\left(\mathbf{s}_{t+1}^m \mid \mathbf{s}_t^m, \mathbb{S}_t\right)\Big|  \nonumber                                               \\
        =& \sum_{\mathbf{s}_{t+1}^c} P\left(\mathbf{s}_{t+1}^c \mid \mathbf{s}_t^c, (\mathbb{S}_t, \mathbb{R}_t)\right) \cdot \sum_{\mathbf{s}_{t+1}^m}\Big|\hat{P}\left(s_{t+1}^m \mid \mathbf{s}_t^m, \mathbb{S}_t\right) \nonumber \\
        & -P\left(\mathbf{s}_{t+1}^m \mid \mathbf{s}_t^m, \mathbb{S}_t\right)\Big|\\
        =& \sum_{\mathbf{s}_{t+1}^c} P\left(\mathbf{s}_{t+1}^c \mid \cdot\right) \cdot\left\|\hat{P}\left(\cdot \mid \mathbf{s}_t^m, \mathbb{S}_t\right)-P\left(\cdot \mid \mathbf{s}_t^m, \mathbb{S}_t\right)\right\|_1  \nonumber   \\
        \overset{(a)}{=} & \left\|\hat{P}\left(\cdot \mid \mathbf{s}_t^m, \mathbb{S}_t\right)-P\left(\cdot \mid \mathbf{s}_t^m, \mathbb{S}_t\right)\right\|_1  \nonumber\\
        \overset{(b)}{=}&2D_{\mathrm{TV}} \left( \hat{P}(\mathbf{s}^{m}_{t+1}|\mathbf{s}^{m}_{t},\mathbb{S}_t) , P(\mathbf{s}^{m}_{t+1}|\mathbf{s}^{m}_{t},\mathbb{S}_t) \right)\overset{(c)}{\leq} 2\varepsilon.\nonumber
    \end{align}

\noindent Specifically, (a) results from isolating the system transition probability as a common factor. (b) is derived by converting the $\ell_1$-norm into TV distance via its standard definition $D_{\mathrm{TV}}(p, q) = \frac{1}{2} \| p-q \|_1$. (c) applies Assumption \ref{as: bounded_p_error} that the TV between the model and true statistical transitions is at most $\varepsilon$. Hence, we have the corollary.
\end{proof}


The following theorem characterizes the bounded deviation of the return-based advantage under model-based dynamics.
\begin{theorem}[Generalized Simulation Lemma for Return-Based Advantage Functions]
    \label{thm: Generalized_Sim}
    Under Assumption \ref{as: bounded_r_error} and Assumption \ref{as: bounded_p_error}, for any fixed policy $\pi$, the difference in cumulative normalized rewards over a finite horizon $K$ between the model-based MDP $\hat{\mathcal{M}}$ and the true environment $\mathcal{M}$ is bounded by
    \begin{align}
        \left| A^\pi_{\hat{P}}(\mathbf{o}, \mathbf{a}) - A^\pi_P(\mathbf{o}, \mathbf{a}) \right| \le  (K^2-K) \tilde{R}_{\max} \cdot \varepsilon.
    \end{align}
\end{theorem}

\begin{proof}
    By definition,
    \begin{align}
             & \left| A_{\hat{P}}^\pi(\mathbf{o}, \mathbf{a})- A_P^\pi(\mathbf{o}, \mathbf{a})\right|\label{eq:def_a_ex}\\
        =    & \left|\mathbb{E}_{\hat{P}, \pi}\left[\sum\nolimits_{j=t}^{K-1} \tilde{r}\left(\mathbf{o}_j, \mathbf{a}_j\right)\right]-\mathbb{E}_{P, \pi}\left[\sum\nolimits_{j=t}^{K-1} \tilde{r}\left(\mathbf{o}_j, \mathbf{a}_j\right)\right]\right| \nonumber \\
        \leq & \sum\nolimits_{t=0}^{K-1}\left|\mathbb{E}_{\hat{P}, \pi}\left[\tilde{r}_t\right]-\mathbb{E}_{P, \pi}\left[\tilde{r}_t\right]\right|.\nonumber
    \end{align}

    Since $\tilde{r}(\mathbf{o}_j, \mathbf{a}_j)$ is bounded by $\tilde{R}_{\max}$ according to  Assumption \ref{as: bounded_r_error}, and by TV Inequality, shown in Prop. 4.2 of \cite{levin2009markov}, we have
    \begin{equation}    \left|\mathbb{E}_{\hat{P},\pi}\left[\tilde{r}_t\right]-\mathbb{E}_{P,\pi}\left[\tilde{r}_t\right]\right| \leq \tilde{R}_{\max } \cdot\left\|\hat{P}^\pi_t-P^\pi_t\right\|_1 \label{eq:tv_ineq},
    \end{equation}
    where $P_t^\pi$ and $\hat{P}_t^\pi$ denote the distributions over $t$-step state-action trajectories induced by policy $\pi$ under transition models $P$ and $\hat{P}$, respectively. The total variation between $K$-step distributions grows at most linearly with the horizon length, as the distributional shift at step $t$ reflects the accumulation of per-step transition errors over the entire trajectory up to time $t$ \cite{kearns2002near}. In other words, by Lemma \ref{c:bounded_p_error}, 
    \begin{align}
        \left\|\hat{P}^\pi_t-P^\pi_t\right\|_1 \leq 2 t \cdot \varepsilon \label{eq:cum_error_p}.
    \end{align}

    By merging \eqref{eq:tv_ineq} and \eqref{eq:cum_error_p} into \eqref{eq:def_a_ex}, we have the theorem after simple mathematical manipulations.
\end{proof}

\noindent
\textit{Remark}: Theorem \ref{thm: Generalized_Sim} result shows that the cumulative advantage error is controllable, growing polynomially with horizon $K$ and linearly with the reward bound and transition error $\varepsilon$. 
Therefore, there are several practical implications. First, the framework is most effective for scenarios with a finite horizon, as the cumulative error may become significant in very long-running tasks. Second, the performance bound is directly tied to the fidelity of the virtual environment. Finally, the reliance on $\tilde{R}_{\max}$ underscores the importance of a well-designed and stable reward function, as volatility or poor scaling can loosen the bound. Fortunately, with
normalized rewards ($R_{\max}\leq 1$) and a Transformer-based transition model, $\varepsilon$ can typically be reduced to a sufficiently accurate level. Together with proper control of $K$, the deviation between model-based and true advantages remains tightly bound. 


\section{Simulation Results}
\label{sec: simulations}
\subsection{Experimental Setup}
\begin{figure*}[!b]
    \centering
    \subfigure[Energy efficiency]{
        \begin{minipage}[t]{0.32\linewidth}
            \centering
            \includegraphics[width=\linewidth]{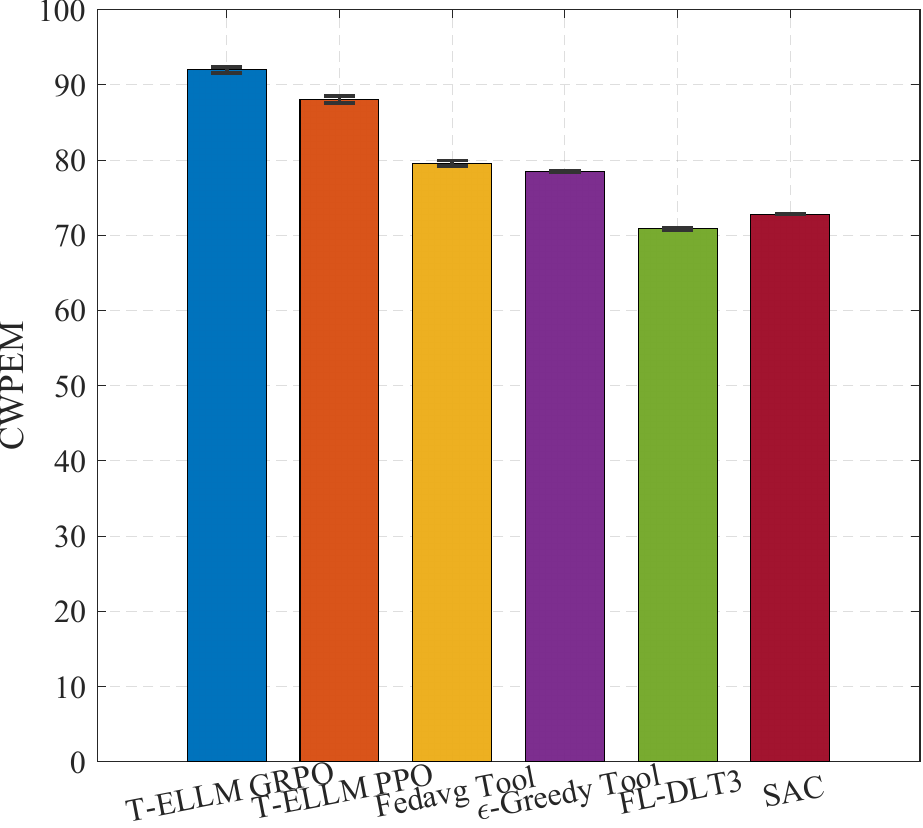}
            \vspace*{-0.5em}
            \label{subfig:energy_enff}
        \end{minipage}}
    \hfill
    \subfigure[Convergence of Wireless FL]{
        \begin{minipage}[t]{0.32\linewidth}
            \centering
            \includegraphics[width=\linewidth]{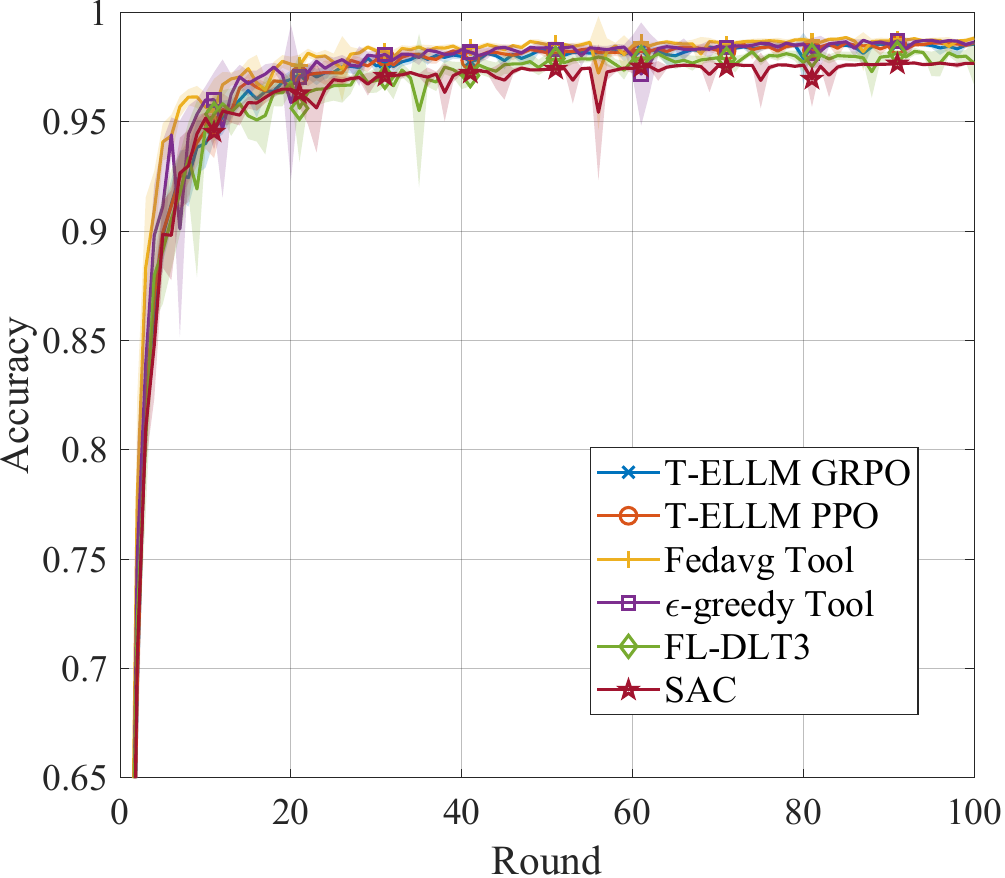}
            \vspace*{-0.5em}
            \label{subfig:acc_round}
        \end{minipage}}
    \hfill
    \subfigure[Energy consumption per round]{
        \begin{minipage}[t]{0.32\linewidth}
            \centering
            \includegraphics[width=\linewidth]{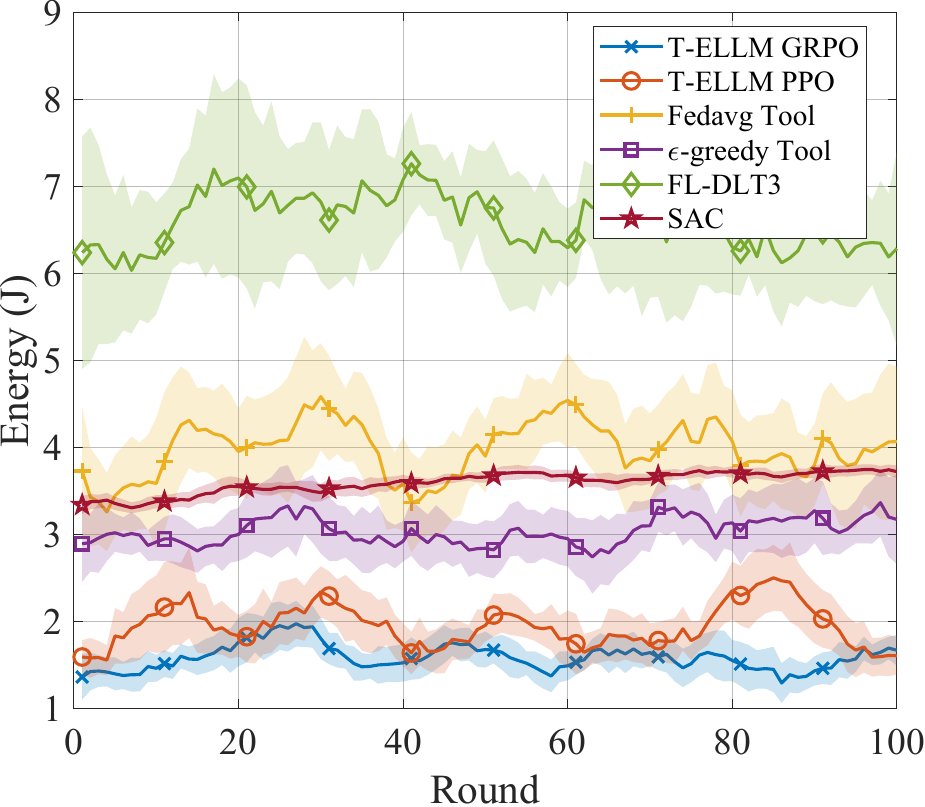}
            \vspace*{-0.5em}
            \label{subfig:energy_per}
        \end{minipage}}
    \vspace{-.5em}
    \caption{Performance comparison of different algorithms for wireless FL.}
    \label{trainenv}
\end{figure*}
\begin{table}[!t]
    \centering
    \caption{Default values of simulation parameters.}
    \label{simulation}
    \begin{tabular}{m{5.5cm}|c}
        \toprule
        \textbf{Parameters}                                                & \textbf{Value}          \\
        \hline
        Total number of devices $N$                                        & $20$                    \\
        \hline
        Number of rounds $K$ (Episodic length)                             & $100$                   \\
        \hline
        Number of local iteration $I$                                      & $5$                     \\
        \hline
        The number of CPU cycles to process one sample $C$                 & $7\times 10^5$          \\
        \hline
        The effective switch capacitance constant $\zeta$                  & $1\times10^{-28}$       \\
        \hline
        The maximum CPU frequency of device $n$ $f_{n,\text{max}}$         & $[0.5, 4.0]$ GHZ        \\
        \hline
        The maximum transmission power of device $n$ $p_{n,\text{max}}$    & $$[0.001, 1]$$ W        \\
        \hline
        The total bandwidth of the system $B$                              & $2$ MHz                  \\
        \hline
        Channel coefficient of device $n$ during the $t$-h round $G_{t,n}$ & $[10^{-7}, 10^{-6}]$ dB \\
        \hline
        The noise power spectral density $N_0$                             & $-174$ dbm/MHz          \\
        \hline
        Weight size of the local model in bits & $53.21$ Mbit            \\
        \hline
        Time-related QoS $T_\text{QoS}$                                                  & $15$ s                  \\
        \hline
        Weight factor $\sigma$                                             & $0.8$                   \\
        \bottomrule
    \end{tabular}
\end{table}

We consider an image classification task in a wireless FL scenario, as discussed in Section \ref{sec: model}, where the total number of devices is $N=20$. In the experiments, we use the well-known MNIST dataset for FL training of a Convolutional Neural Network (CNN) model with the cross-entropy loss function.
To simulate data heterogeneity across devices, we sample label ratios and dataset sizes from a Dirichlet distribution parameterized by $\alpha$, which controls the degree of non-IIDness. Notably, a smaller $\alpha$ leads to more non-IID data, while a larger $\alpha$ results in more homogeneous data.  We set $\alpha=0.2$ as the default to simulate the non-IID data distribution. Other default configurations of simulation parameters are specified in the Table \ref{simulation}.

For the training of T-ELLM, we employ the ALTD  \cite{9187798} as a complementary tool for dynamic management of CPU frequency and transmission power, as well as equal, fixed bandwidth allocation. In addition, the environmental model, which is adapted from the decoder-only part of the GPT-2-small architecture \cite{Radford2018ImprovingLU, NEURIPS2021_7f489f64} with two extra linear layers, is trained first. Subsequently, a LLaMA-3-1B-based \cite{grattafiori2024llama3herdmodels} T-ELLM is used for GRPO \cite{shao2024deepseekmathpushinglimitsmathematical}-based policy learning and generation. Besides, to evaluate the performance of the proposed T-ELLM, we compare it with the following baselines.
\begin{itemize}
    \item FedAvg Tool\cite{mcmahan2017communication}: A specific proportion of clients are randomly selected to participate in each round of FL training. Since it does not have the function of resource allocation itself, we use the ALTD tool to make resource allocation decisions.
    \item $\epsilon$-Greedy Tool\cite{273723}: A variant of the Oort algorithm. Specifically, energy considerations are added, and tools are leveraged to obtain time and energy reference. Moreover, we assume all the device time and energy consumption can be known in advance, so that the $\epsilon$-greedy algorithm can be executed \cite{273723}.
    \item FL-DLT3 \cite{9779339}: FL-DLT3 enables a twin-delayed deep deterministic policy gradient (TD3) framework to optimize accuracy and energy balance in a continuous domain. Compared to transmission power allocation for FL efficiency optimization \cite{9779339}, we further expand it to manage CPU frequency with equal, fixed bandwidth allocation.
    \item SAC (Soft Actor-Critic)\cite{10368103,pmlr-v80-haarnoja18b}: SAC is an off-policy actor-critic deep RL algorithm based on the maximum entropy framework\cite{pmlr-v80-haarnoja18b}. \cite{10368103} applies SAC to wireless FL scenarios for resource allocation. We implement SAC on top of a transformer decoder-only neural network.
    \item T-ELLM PPO: It uses PPO \cite{schulman_proximal_2017} rather than GRPO \cite{shao2024deepseekmathpushinglimitsmathematical} to fine-tune T-ELLM. Compared to GRPO, PPO employs an additional critic network to optimize the NN.
\end{itemize}

\subsection{Performance Comparison}
\label{sec:per_com}
\begin{figure}[!t]
    \begin{center}
        \vspace*{-1.5em}{
            \includegraphics[width=.475\textwidth]{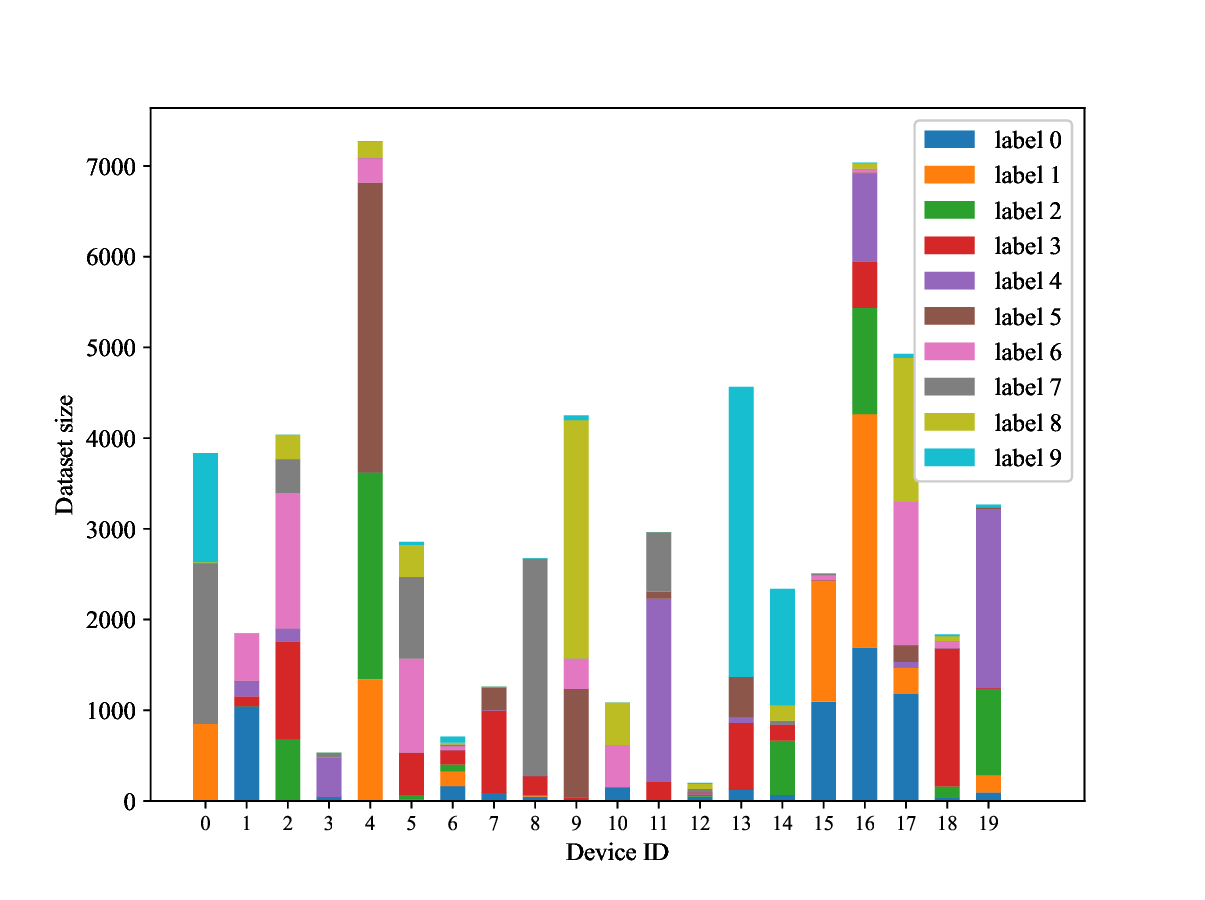}}
    \end{center}
    \vspace*{-1em}
    \caption{Label and dataset size distributions for non-IID data on different devices.}\label{simu_noniid}\vspace{-0.5em}
\end{figure}
\begin{figure}[tp]
    \begin{center}
        \vspace*{-0.5em}{
            \includegraphics[width=.4\textwidth]{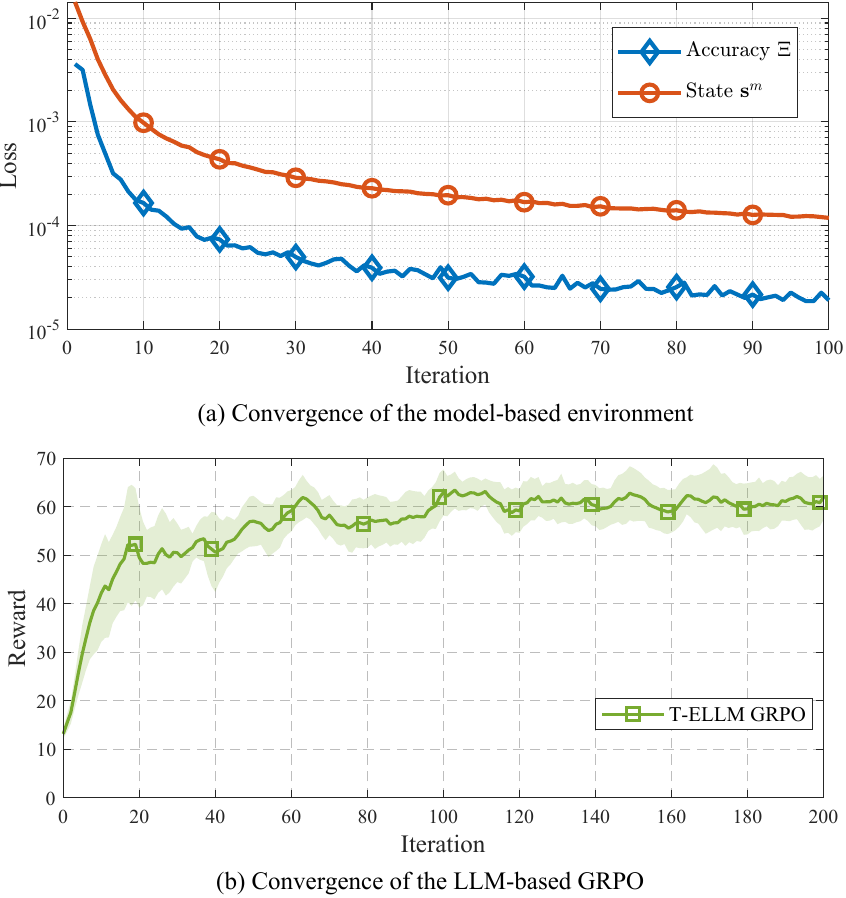}}
    \end{center}
    \vspace*{-1em}
    \caption{{Convergence of the model-based virtual environment and GRPO-based T-ELLM.}}\label{conv}\vspace{-1em}
\end{figure}
We first show the overall performance comparison, in terms of CWEPM in \eqref{p1a}, learning accuracy, and energy consumption per round, and Fig. \ref{trainenv} presents the corresponding results. Notably, as shown in Fig. \ref{simu_noniid}, significantly heterogeneous label distributions and dataset sizes exist for evaluation. As shown in Fig. \ref{subfig:energy_enff}, the proposed T-ELLM GRPO and T-ELLM PPO achieve the highest CWPEM values, suggesting that these methods require less energy to attain target performance levels. Fig. \ref{subfig:acc_round} illustrates the convergence of wireless FL under different algorithms. The proposed T-ELLM GRPO and T-ELLM PPO exhibit a remarkable convergence rate towards the desired accuracy.
Additionally, Fig. \ref{subfig:energy_per} presents the energy consumption per round of wireless FL, where the solid curves therein represent the mean testing accuracy across $10$ experimental trials and the shaded part represents the $95\%$ confidence interval calculated from these experiments (the other figures are likewise). The proposed T-ELLM GRPO and T-ELLM PPO consume less energy per round, demonstrating their efficiency in resource utilization during the FL process. Furthermore, the narrower confidence interval observed for T-ELLM GRPO and T-ELLM PPO, relative to the FL-DLT3 and Fedavg Tool baseline, indicates stable and consistent decision-making, while maintaining a level of stability comparable to the heuristic methods.

To implement the proposed T-ELLM, $600$ trajectories of $((\mathbf{s}^{m}_t, \varXi(\mathbb{S}_{t-1}), \mathbb{S}_{t}),(\mathbf{s}^{m}_{t+1}, \varXi(\mathbb{S}_{t})))$ are collected to train the GPT model as the statistics part of the virtual environment.
The convergence of the model-based environment is shown in Fig. \ref{conv}(a). It can be observed from Fig. \ref{conv}(a) that the MSE loss for both state $\mathbf{s}^m_t$ and accuracy $\varXi(\mathbb{S}_t)$ decreases steadily. While the accuracy metric consists of a single value, thus being relatively easier to predict, the state encompasses more complex information. Nevertheless, the state also converges to a satisfactory result. Based on the trained environmental model, the evolution of T-ELLM can be carried out offline. Fig. \ref{conv}(b) presents the reward progression during GRPO fine-tuning. It can be seen that under the optimization of GRPO, the reward value gradually stabilizes at a higher level, which indicates the satisfactory convergence of T-ELLM.


\subsection{Generalizable Capability to Environmental Changes}
\begin{figure}[tp]
    \begin{center}
        \vspace*{-1em}{
            \includegraphics[width=.425\textwidth]{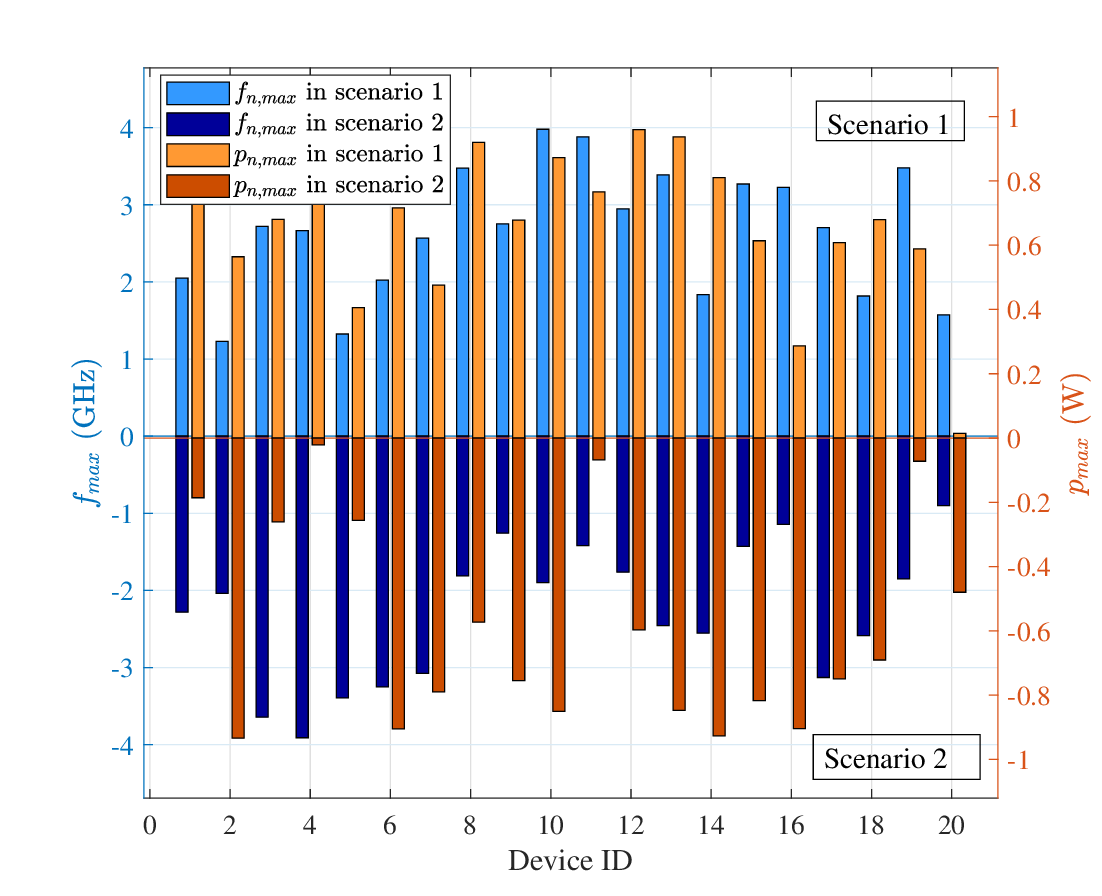}}
    \end{center}
    \vspace*{-0.5em}
    \caption{Comparison of changing environment with respect to $f_{n,\text{max}}$ and $p_{n,\text{max}}$.}\label{changpf}\vspace{-0.5em}
\end{figure}
\begin{figure*}[!t]
    \centering
    \subfigure[Energy efficiency]{
        \begin{minipage}[t]{0.32\linewidth}
            \centering
            \includegraphics[width=\linewidth]{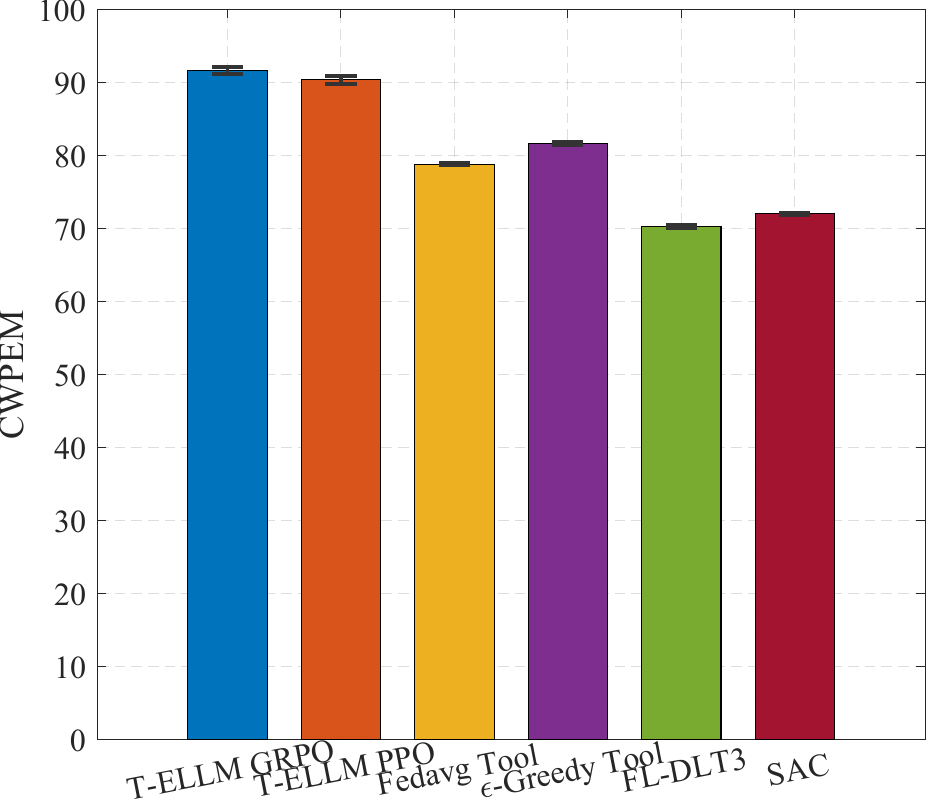}
            \label{env_energy_per}
        \end{minipage}}
    \hfill
    \subfigure[Convergence of Wireless FL]{
        \begin{minipage}[t]{0.32\linewidth}
            \centering
            \includegraphics[width=\linewidth]{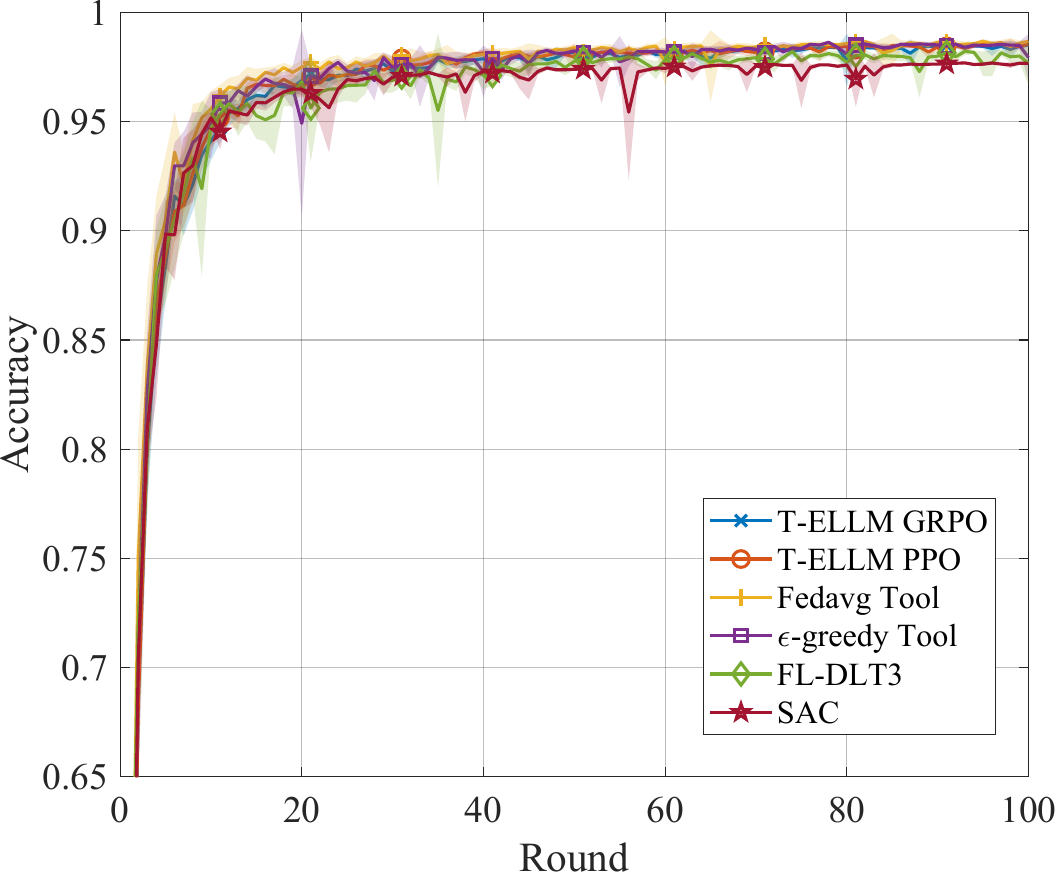}
            \label{env_acc_per}
        \end{minipage}}
    \hfill
    \subfigure[Energy consumption per round]{
        \begin{minipage}[t]{0.32\linewidth}
            \centering
            \includegraphics[width=\linewidth]{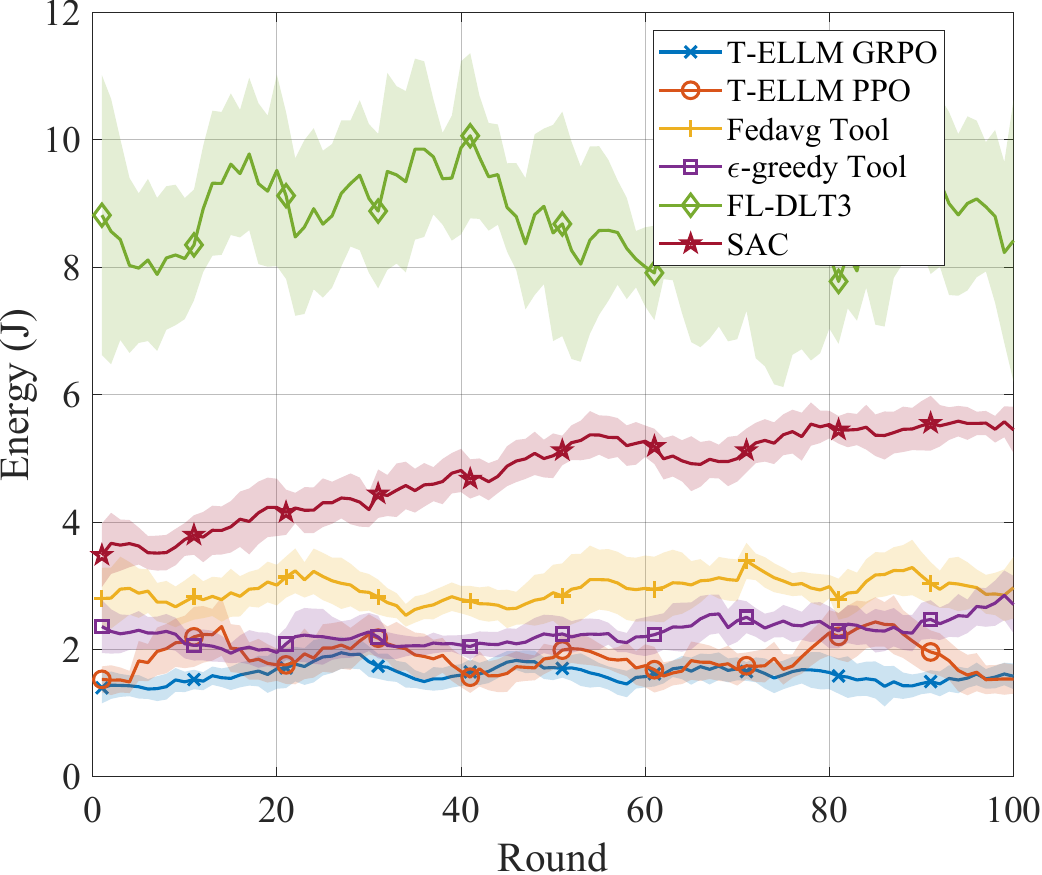}
            \label{subfig:env_energy_per}
        \end{minipage}}
    \vspace{-0.3cm}
    \caption{Performance comparison of different algorithms for wireless FL with Scenario 2.}
    \label{fpchange}
    \vspace{-0.1cm}
\end{figure*}
\subsubsection{Adaptation to Resource Changes} 
To verify the generalization of the proposed T-ELLM to the environment, we first change the computing and communication capabilities of each device in the environment. Fig. \ref{changpf} highlights the differences in terms of the maximum CPU frequency and transmission power of devices. Compared to the environment (i.e., Scenario 1) for evaluation in Section \ref{sec:per_com}, the modified environment is termed Scenario 2. Notably, all algorithms are directly transferred from the training outcome in Scenario 1 and have no prior interaction with Scenario 2, ensuring an unbiased assessment of adaptability. The corresponding performance in Scenario 2 is shown in Fig. \ref{fpchange}. As demonstrated in Fig. \ref{env_energy_per}, the CWPEM of T-ELLM GRPO and T-ELLM PPO outperforms that of other baseline algorithms, indicating superior performance with lower energy consumption. Furthermore, Fig. \ref{env_acc_per} confirms that the learning accuracy of T-ELLM remains robust, without sacrificing the learning efficiency.
Furthermore, Fig. \ref{subfig:env_energy_per} shows that the energy consumption per round yielded by T-ELLM is also lower than that of other algorithms, demonstrating its ability to maintain high accuracy while optimizing energy efficiency. These results highlight T-ELLM's adaptability to varying environmental conditions.
Additionally, the FedAvg tool and $\epsilon$-Greedy tool are not learning-based algorithms, naturally, their performance is less affected by environmental changes. In comparison, as shown in Fig. \ref{subfig:env_energy_per}, due to its training only in the default Scenario 1, the energy consumption of FL-DLT3 and SAC is significantly higher than that of other algorithms. This further underscores the advantages of the proposed T-ELLM in terms of environmental adaptability and energy efficiency.

\begin{figure*}[tp]
    \centering
    \subfigure[Energy efficiency]{
        \begin{minipage}[t]{0.32\linewidth}
            \centering
            \includegraphics[width=\linewidth]{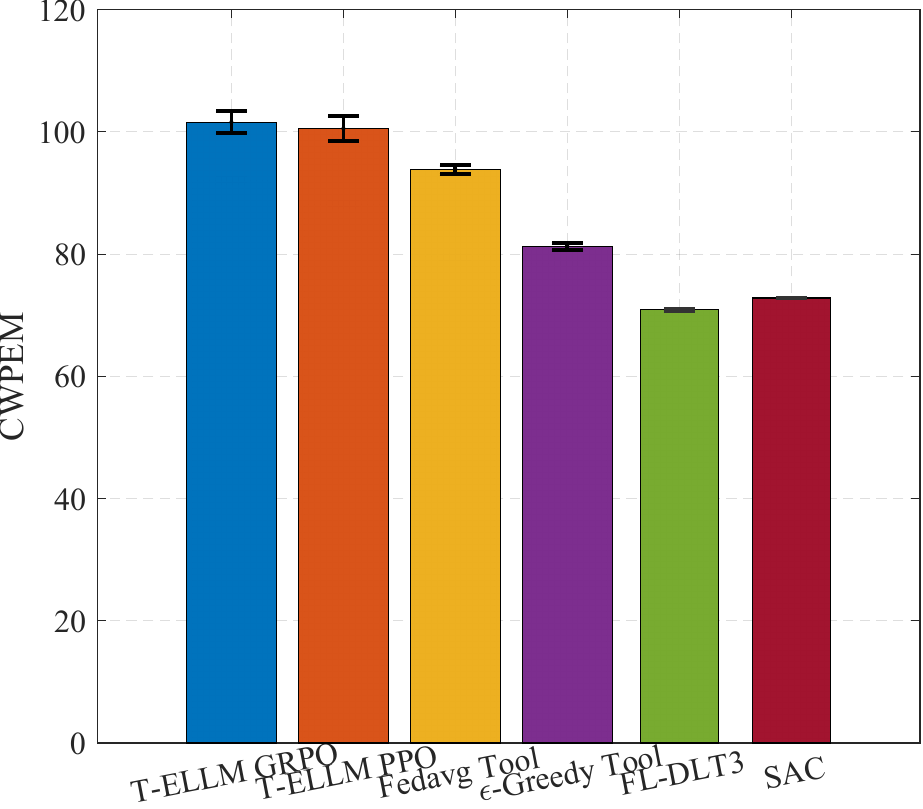}
            \label{qos_energy_eff}
        \end{minipage}}
    \hfill
    \subfigure[Convergence of Wireless FL]{
        \begin{minipage}[t]{0.32\linewidth}
            \centering
            \includegraphics[width=\linewidth]{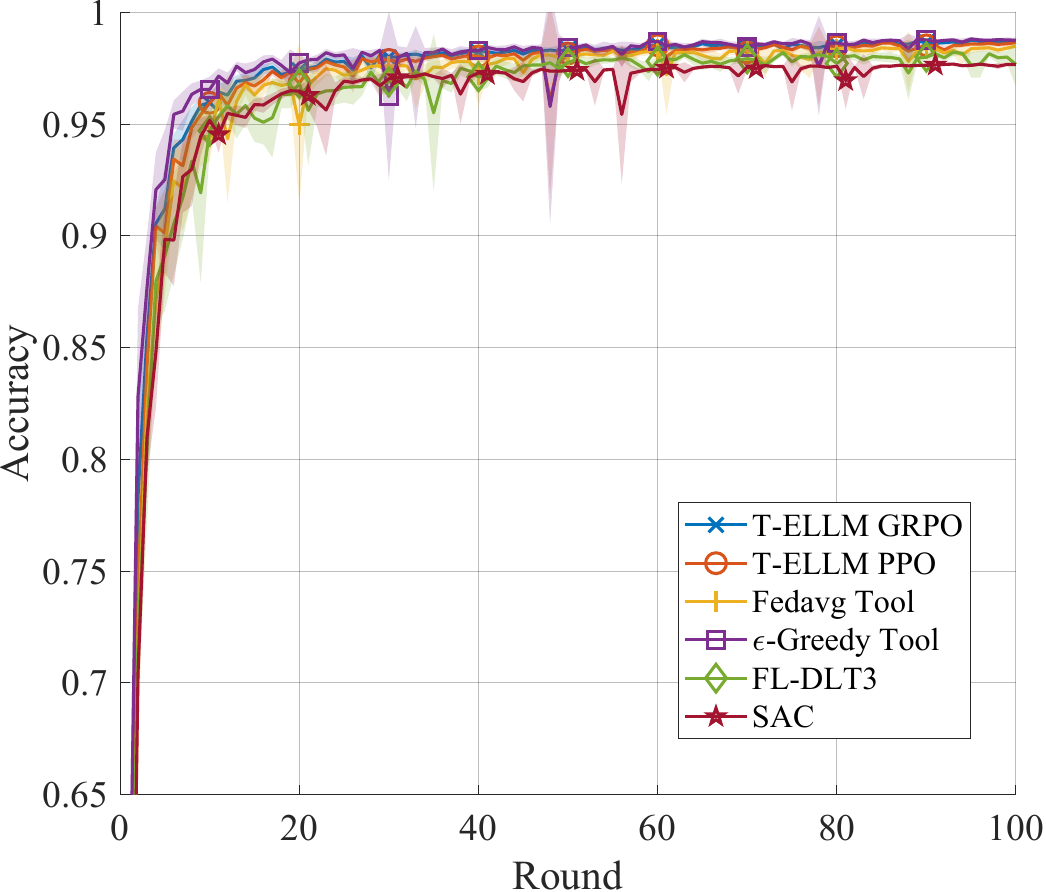}
            \label{qos_acc_per}
        \end{minipage}}
    \hfill
    \subfigure[Energy consumption per round]{
        \begin{minipage}[t]{0.32\linewidth}
            \centering
            \includegraphics[width=\linewidth]{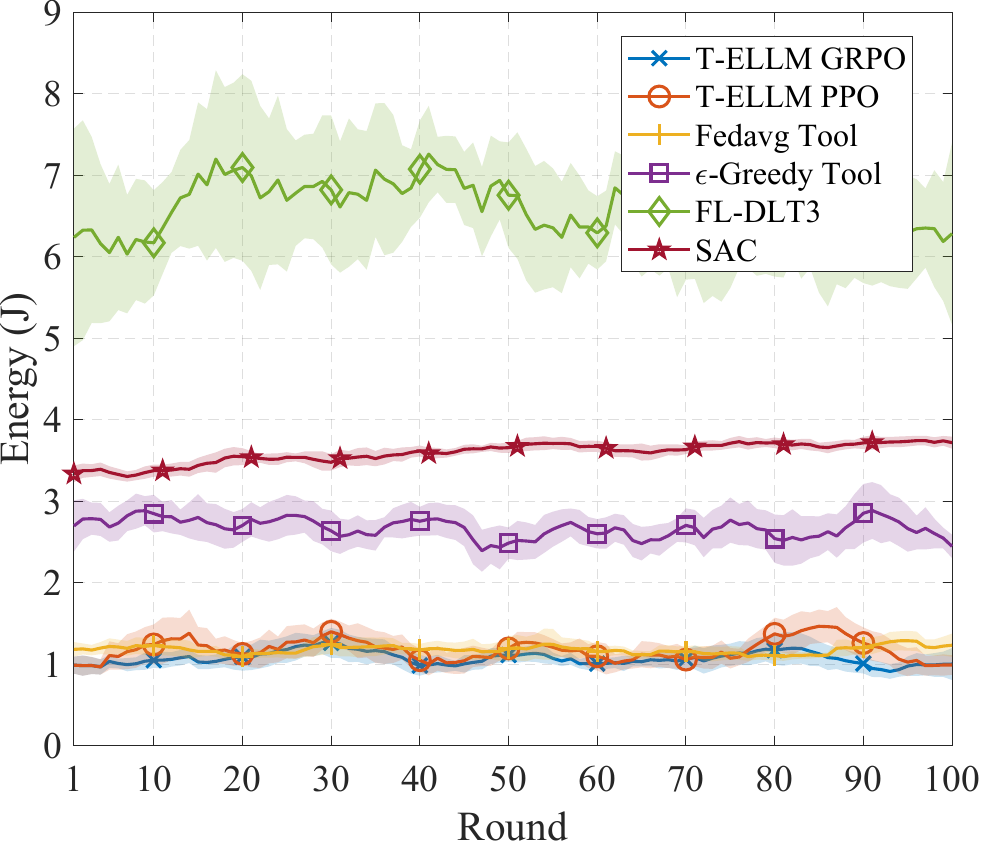}
            \label{qos_energy_per}
        \end{minipage}}
    \vspace{-0.3cm}
    \caption{Performance comparison of different algorithms for wireless FL with $T_\text{QoS} = 20$ s.}
    \vspace{-0.1cm}
    \label{qoschange}
\end{figure*}

\begin{figure}[tp]
    \begin{center}
        \vspace*{-0.5em}{
            \includegraphics[width=.425\textwidth]{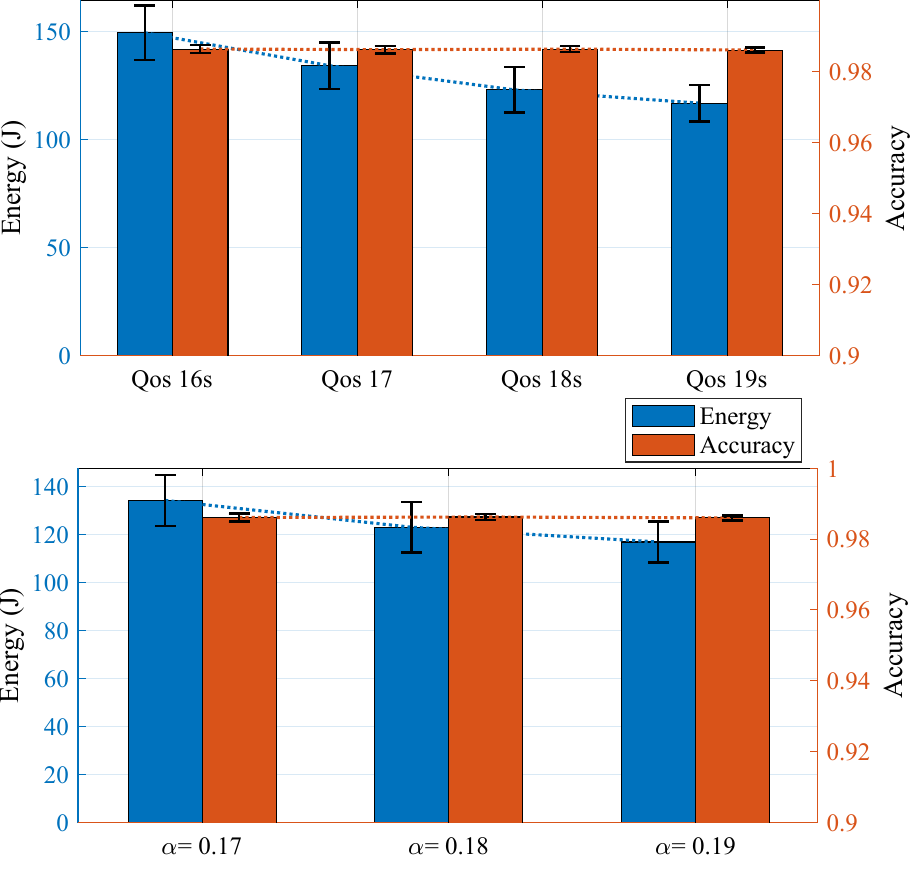}}
    \end{center}
    \vspace*{-0.5em}
    \caption{Performance of the proposed T-ELLM GRPO with different QoS requirements and data heterogeneity.}\label{changing}\vspace{-0.5em}
\end{figure}
\subsubsection{Adaptation to Changed Task Requirements}
In this part, we further demonstrate the ability of T-ELLM to cope with different QoS requirements. Fig. \ref{qoschange} provides the results after changing $T_\text{QoS}$ in the default Scenario 1 from $15$ seconds to $20$ seconds.
The results show that the proposed T-ELLM GRPO and T-ELLM PPO achieve the highest CWPEM values in Fig. \ref{qos_energy_eff}, indicating superior energy efficiency in wireless FL. 
Fig. \ref{qos_acc_per} illustrates the convergence of wireless FL under different algorithms. The proposed T-ELLM GRPO and T-ELLM PPO exhibit a significant convergence rate, achieving the desired accuracy. Fig. \ref{qos_energy_per} shows that the proposed T-ELLM GRPO and T-ELLM PPO consume less energy per round, while it can be seen that the FedAvg Tool also maintains a low energy consumption per round. The reason lies in that the resource tool is used therein for resource allocation, contributing to saving overall energy consumption required for the anticipated QoS. On the other hand, although the $\epsilon$-Greedy tool algorithm also gets help from the tool, the extensive reliance on the artificial setting of greedy strategy parameters undermines the potential benefit. Additionally, as shown in  Fig. \ref{qos_energy_per},  FL-DLT3 and SAC demonstrate notably higher energy consumption, as they depend solely on their NNs for resource allocation and lack adaptability to varying QoS demands. In comparison, the proposed T-ELLM framework dynamically adjusts to QoS requirements, enabling more efficient decision-making.

We also evaluate the performance of the proposed T-ELLM with different QoS requirements and heterogeneity. The performance of the proposed T-ELLM is shown in Fig. \ref{changing}. It can be seen in Fig. \ref{changing}(a) that the proposed T-ELLM maintains comparable accuracy with different QoS requirements. Furthermore, the energy consumption decreases with the increase of QoS requirements. Fig. \ref{changing}(b) shows the performance of the proposed T-ELLM with different levels of data heterogeneity, and similar observations can be attained. These results show that the proposed T-ELLM can adapt to environmental changes by yielding appealing accuracy.

\subsection{Performance of TELLM in Different Model Sizes}

\begin{figure*}[tp]
    \centering
    \subfigure[Energy efficiency]{
        \begin{minipage}[t]{0.32\linewidth}
            \centering
            \includegraphics[width=\linewidth]{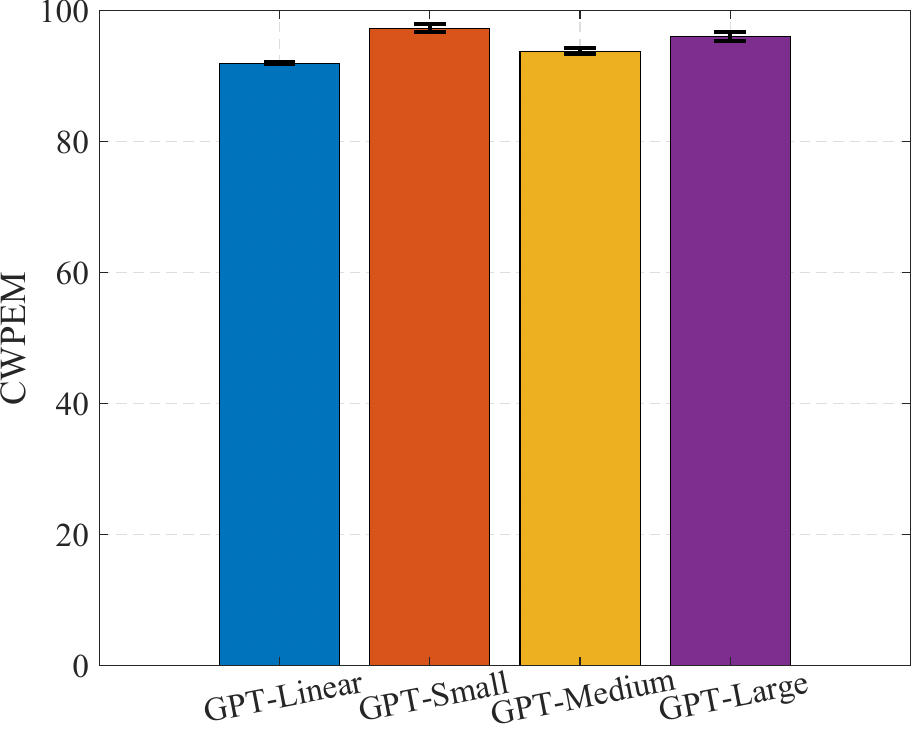}
            \label{gpt_energy_eff}
        \end{minipage}}
    \hfill
    \subfigure[Convergence of Wireless FL]{
        \begin{minipage}[t]{0.32\linewidth}
            \centering
            \includegraphics[width=\linewidth]{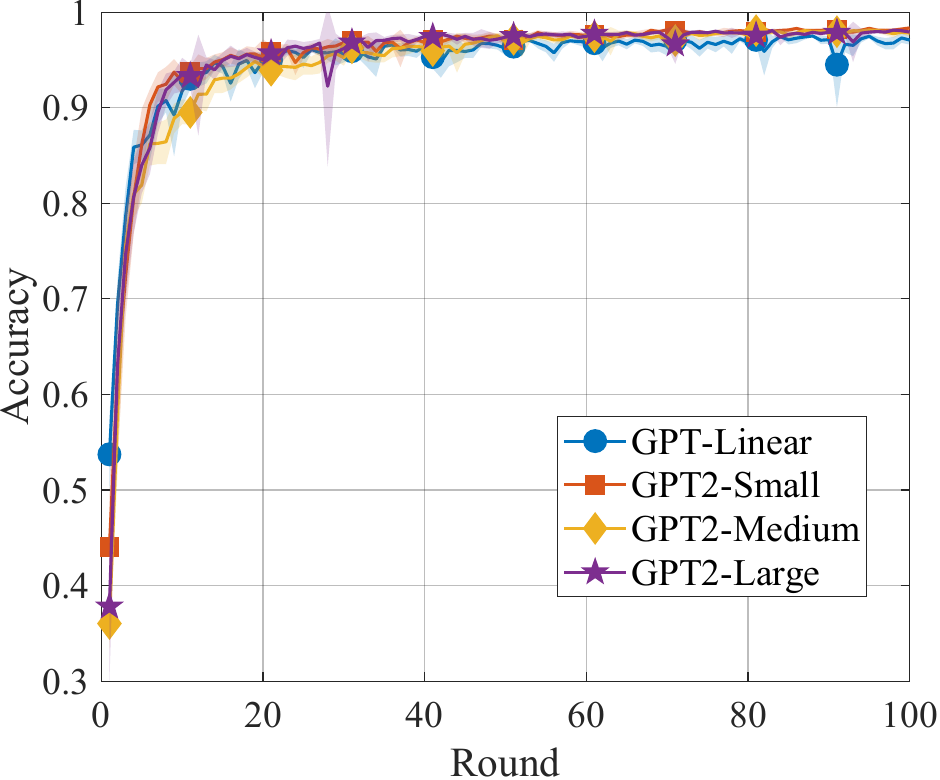}
            \label{gpt_acc_per}
        \end{minipage}}
    \hfill
    \subfigure[Energy consumption per round]{
        \begin{minipage}[t]{0.32\linewidth}
            \centering
            \includegraphics[width=\linewidth]{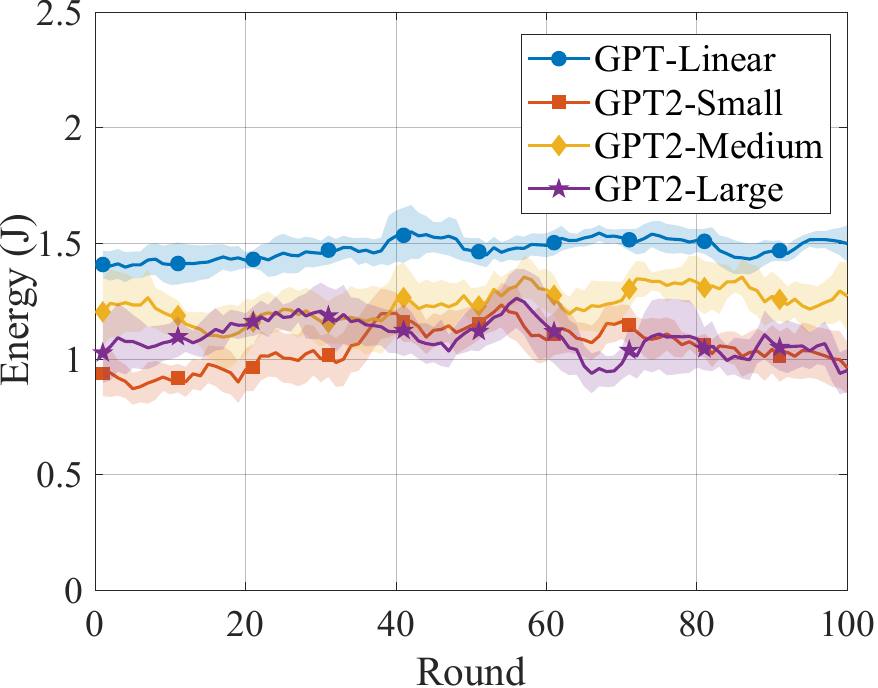}
            \label{gpt_energy_per}
        \end{minipage}}
    \vspace{-0.3cm}
    \caption{Performance comparison of large models with different scales for wireless FL.}
    \vspace{-0.5cm}
    \label{gptmodel}
\end{figure*}

In order to validate the effect of LLM size and language training approach on the T-ELLM, we evaluate the performance of three versions of GPT-2, including GPT-2 Small, GPT-2 Medium, and GPT-2 Large, corresponding to parameter scales of $0.1$B, $0.3$B, and $0.7$B, respectively. Additionally, we compare the performance of GPT-2 models against a more lightweight, task-specific baseline we call GPT-Linear, which uses a Transformer decoder-only architecture (like GPT-2). The result is shown in Fig. \ref{gptmodel}. It can be seen that the performance for the GPT-2 series is all very similar, with no significant difference. Compared with the GPT-2 series, the CWPEM and convergence results of GPT-Linear are slightly lower, while the confidence interval of GPT-Linear is narrower than that of the GPT-2 series.
These results validate the advantage of adopting a linguistic model, and a computationally efficient GPT-2 Small model already yields satisfactory results.

\subsection{The Adaptability of TELLM to Different FL Algorithms}

We investigate the adaptability and robustness of our T-ELLM framework for FL with gradient quantization\cite{10368103} and Differential Privacy (DP)\cite{10.1145/3625558}.
\subsubsection{Adaptation to Gradient Quantization} 
\begin{figure*}[tp]
    \centering
    \subfigure[Energy and accuracy]{
        \begin{minipage}[t]{0.31\linewidth}
            \centering
            \includegraphics[width=\linewidth]{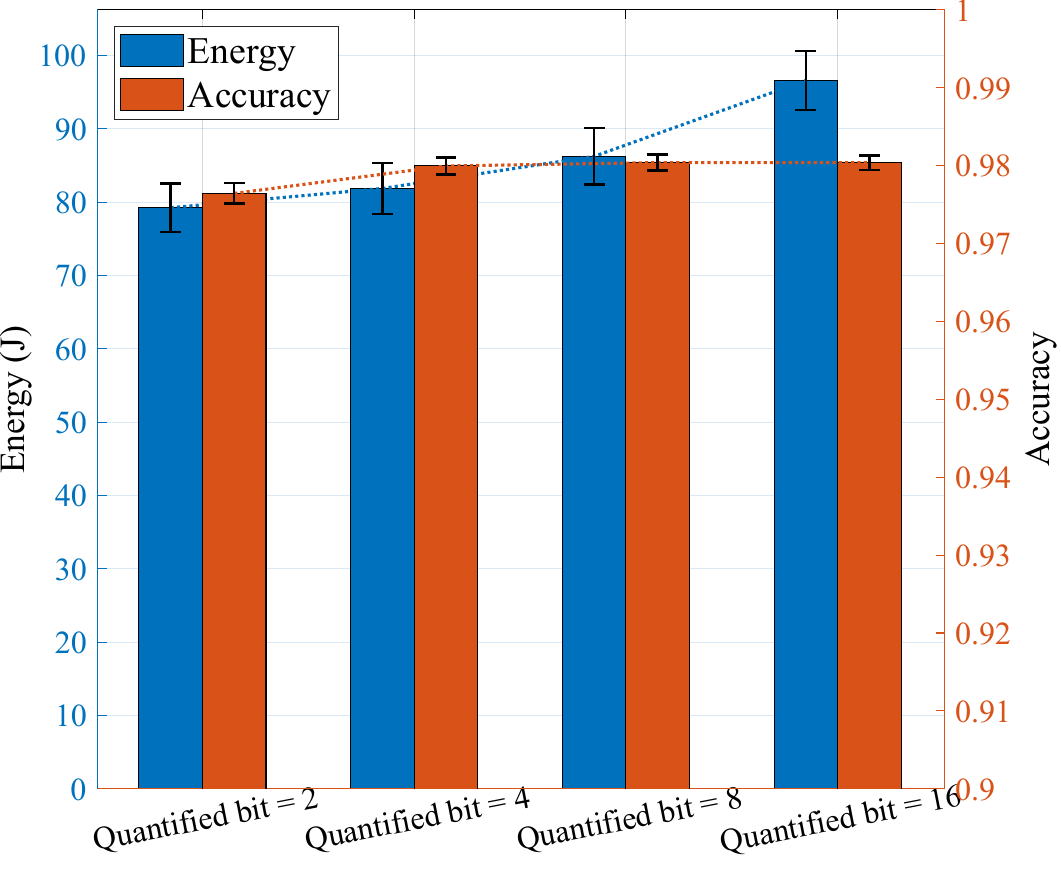}
            \label{gq_en_ac}
        \end{minipage}}
    \hfill
    \subfigure[Convergence of Wireless FL]{
        \begin{minipage}[t]{0.31\linewidth}
            \centering
            \includegraphics[width=\linewidth]{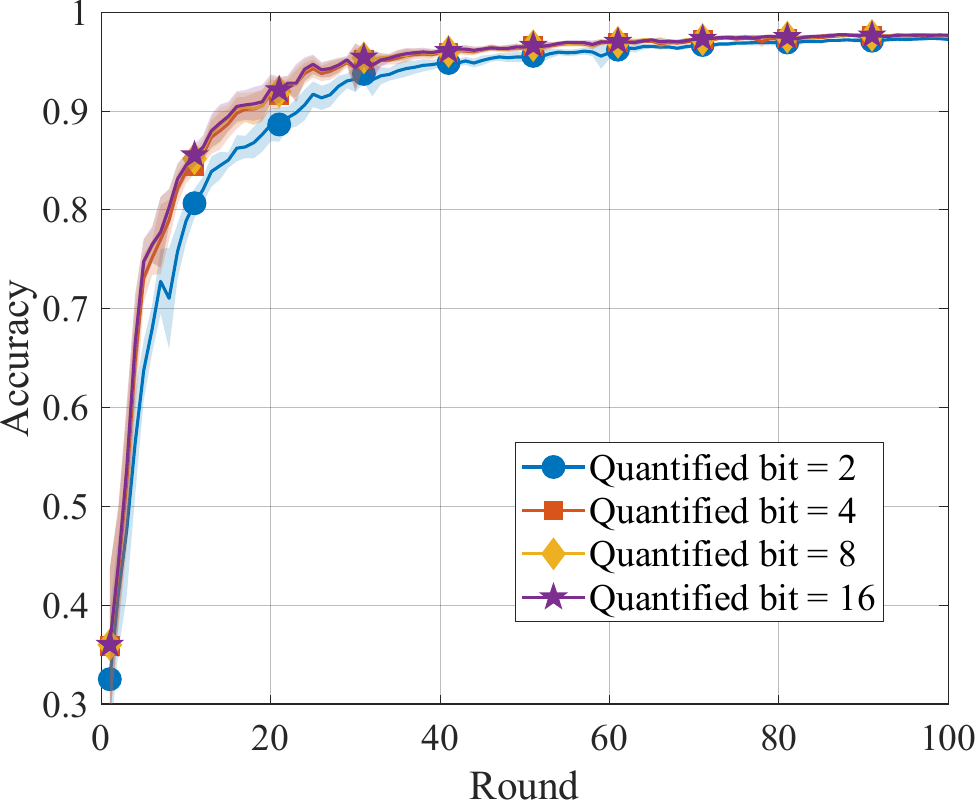}
            \label{gq_acc_round}
        \end{minipage}}
    \hfill
    \subfigure[Energy consumption per round]{
        \begin{minipage}[t]{0.31\linewidth}
            \centering
            \includegraphics[width=\linewidth]{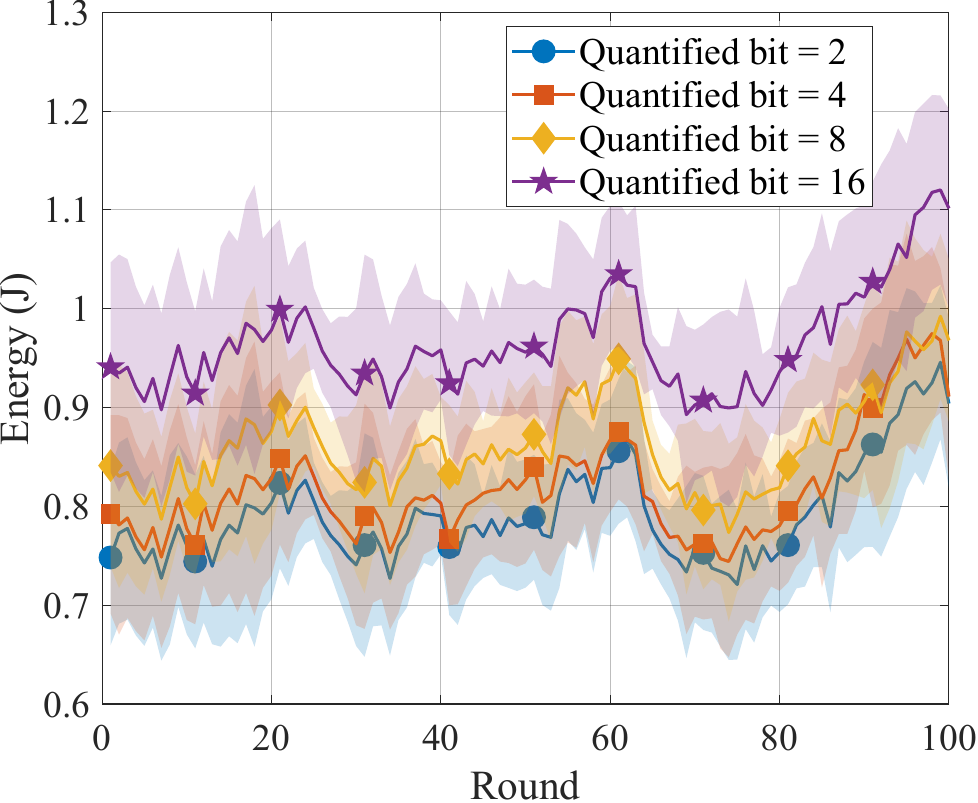}
            \label{gq_energy_per}
        \end{minipage}}
    \vspace{-0.3cm}
    \caption{Performance of the T-ELLM algorithm for wireless FL using gradient quantization.}
    \vspace{-0.5cm}
    \label{gqfl}
\end{figure*}

We present the performance of T-ELLM with varying levels of gradient quantization in Fig. \ref{gqfl}. As shown in Fig. \ref{gq_en_ac}, consistent with our intuition, the energy efficiency of T-ELLM decreases as the number of quantization bits decreases, which is expected since lower bit quantization introduces more noise, potentially degrading model accuracy. However, even with $2$-bit quantization, T-ELLM maintains a reasonable balance between energy consumption and accuracy. Fig. \ref{gq_acc_round} shows that T-ELLM converges to a satisfactory accuracy level across all quantization levels, although higher bit quantization leads to faster convergence and higher final accuracy. Fig. \ref{gq_energy_per} indicates that energy consumption per round increases with the number of quantization bits, as higher precision requires more data transmission. Overall, T-ELLM demonstrates robust performance even under aggressive gradient quantization. 

\subsubsection{Adaptation to Differential Privacy} 
\begin{figure*}[tp]
    \centering
    \subfigure[Energy and accuracy]{
        \begin{minipage}[t]{0.32\linewidth}
            \centering
            \includegraphics[width=\linewidth]{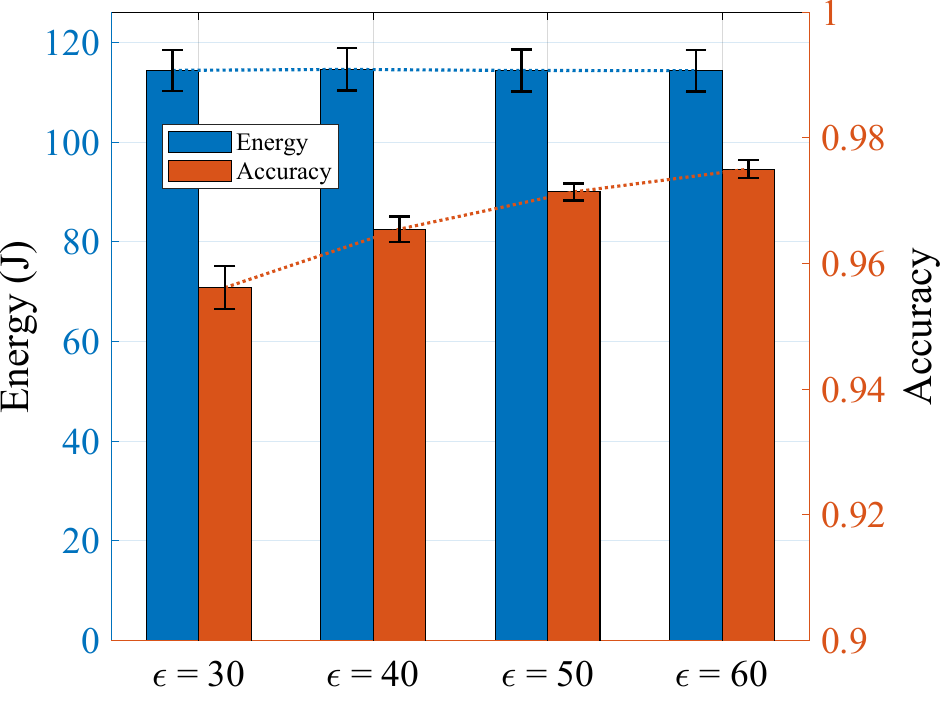}
            \label{dp_en_ac}
        \end{minipage}}
    \hfill
    \subfigure[Convergence of Wireless FL]{
        \begin{minipage}[t]{0.32\linewidth}
            \centering
            \includegraphics[width=\linewidth]{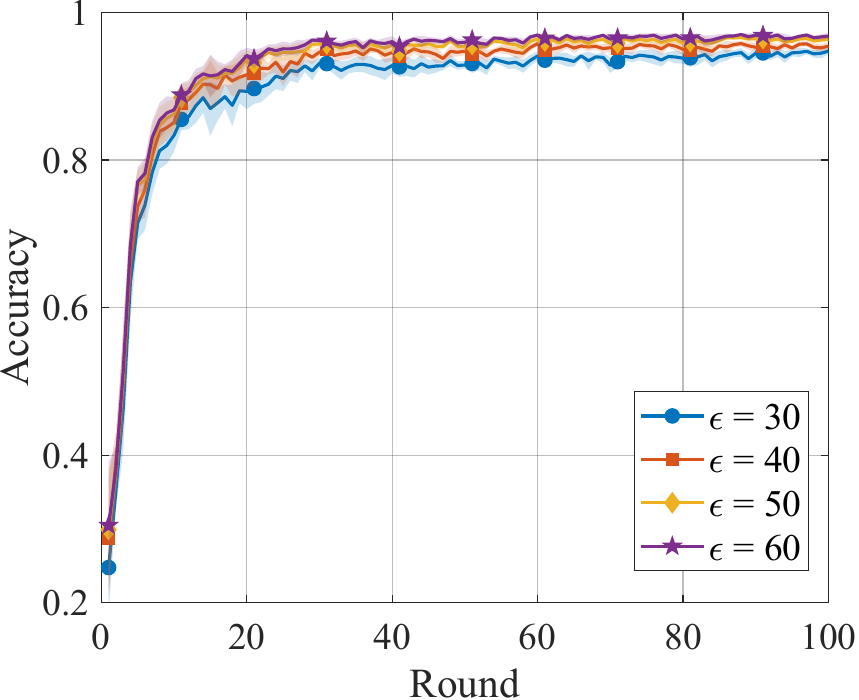}
            \label{dp_acc_round}
        \end{minipage}}
    \hfill
    \subfigure[Energy consumption per round]{
        \begin{minipage}[t]{0.32\linewidth}
            \centering
            \includegraphics[width=\linewidth]{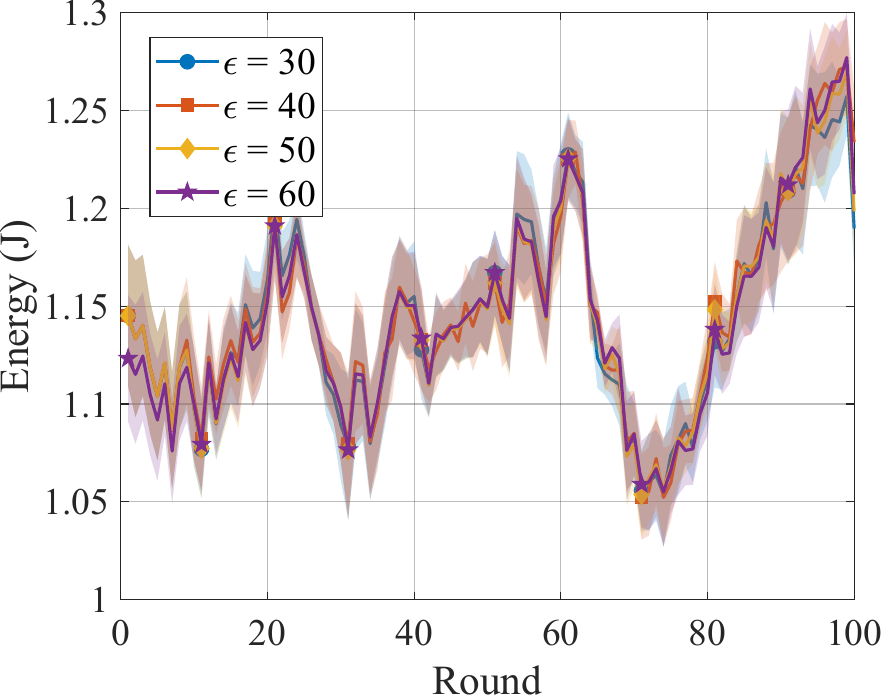}
            \label{dp_energy_per}
        \end{minipage}}
    \vspace{-0.3cm}
    \caption{Performance of the T-ELLM algorithm for wireless FL using differential privacy.}
    \vspace{-0.5cm}
    \label{dpfl}
\end{figure*}

We implement $(\epsilon, \delta)$-DP in the wireless FL process, where $\delta$ accounts for the probability that plain $\epsilon$-DP is broken, and $\epsilon$ controls the level of privacy, where a smaller $\epsilon$ signifies stronger privacy (and more noise). 
We evaluate the performance with various privacy budgets.  Fig.\ref{dp_en_ac} shows that stronger privacy (a lower $\epsilon$) leads to lower final model accuracy, as the increased noise makes learning more difficult. 
Fig. \ref{dp_acc_round} demonstrates that while all settings eventually reach a stable accuracy, stronger privacy constraints slow down the learning process. Fig. \ref{dp_energy_per} illustrates that energy consumption is not significantly affected by the DP level. Overall, T-ELLM maintains robust performance when differential privacy is applied.

\subsection{Scalability for Large-Scale Scenario}

\begin{figure*}[tp]
    \centering
    \subfigure[Energy efficiency]{
        \begin{minipage}[t]{0.32\linewidth}
            \centering
            \includegraphics[width=\linewidth]{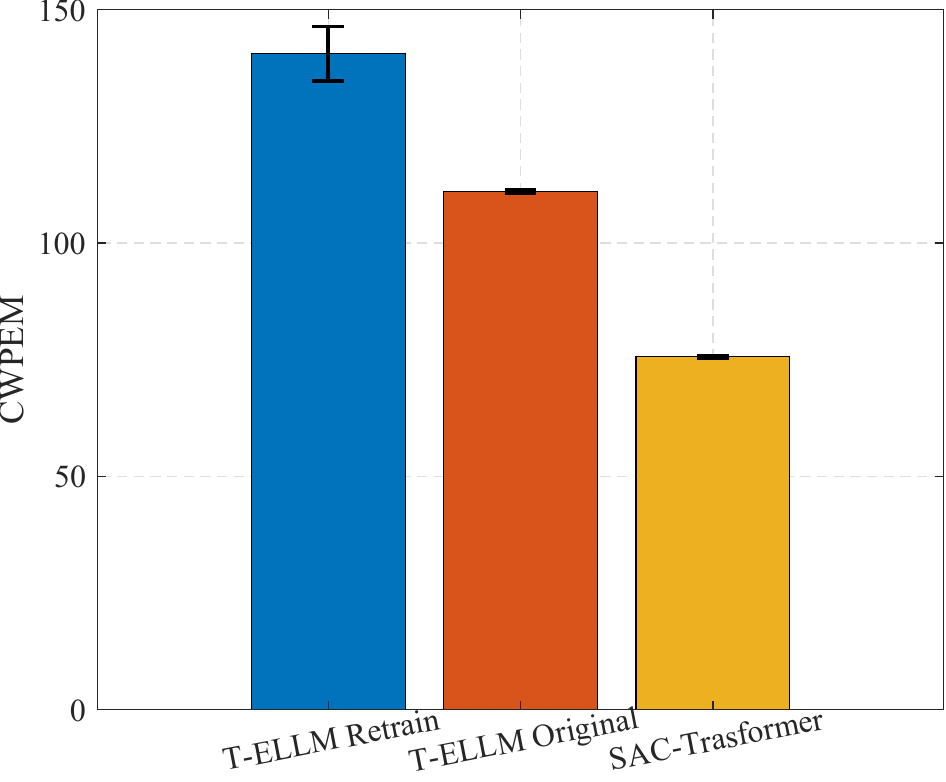}
            \label{sca_energy_eff}
        \end{minipage}}
    \hfill
    \subfigure[Convergence of Wireless FL]{
        \begin{minipage}[t]{0.32\linewidth}
            \centering
            \includegraphics[width=\linewidth]{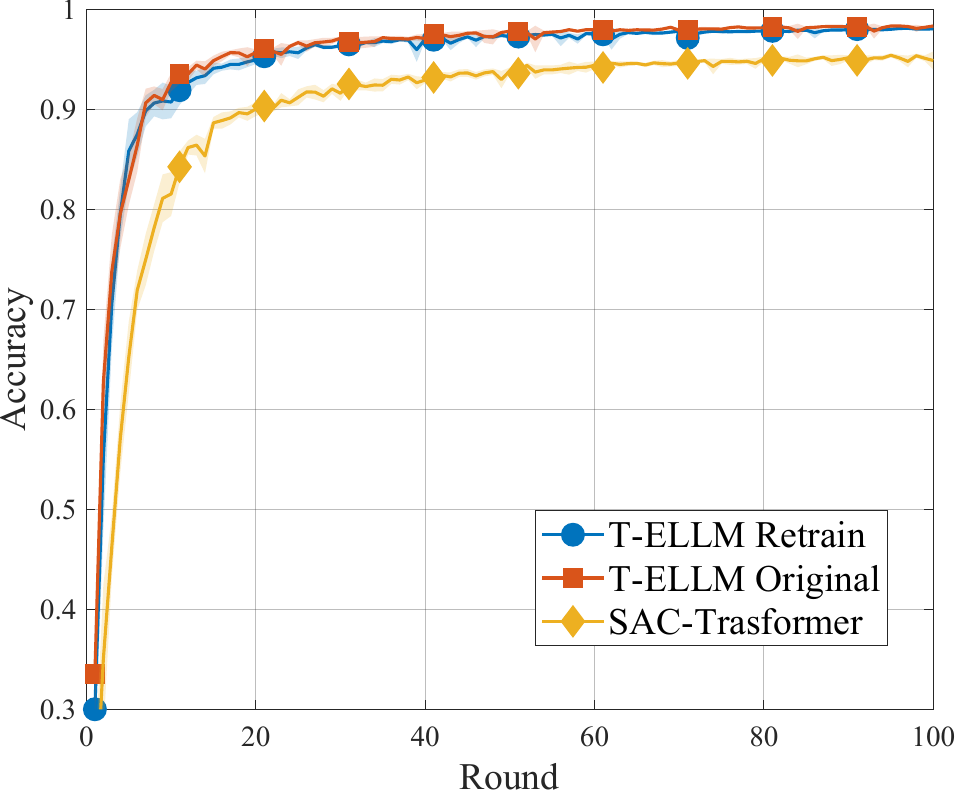}
            \label{sca_acc_per}
        \end{minipage}}
    \hfill
    \subfigure[Energy consumption per round]{
        \begin{minipage}[t]{0.32\linewidth}
            \centering
            \includegraphics[width=\linewidth]{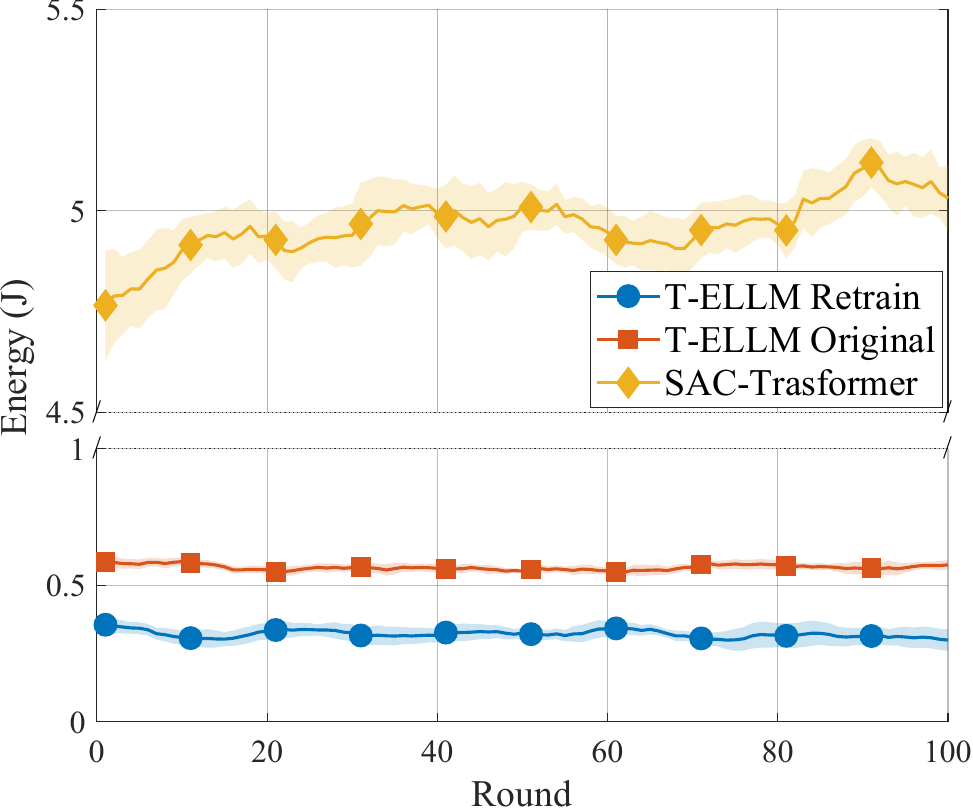}
            \label{sca_energy_per}
        \end{minipage}}
    \vspace{-0.3cm}
        \caption{Performance comparison of learning-based algorithms migrated to the scenario with $100$ devices.}
    \vspace{-0.5cm}
    \label{scamodel}
\end{figure*}

To evaluate the scalability of T-ELLM, we simulate a larger weirless FL environment consisting of $100$ devices, where we distribute the CIFAR-10 dataset across these devices and configure the task to select $10$ participating devices in each training round. 
Meanwhile, we compare the performance of three algorithms:
\begin{itemize}
    \item T-ELLM Retrain: The T-ELLM agent is retrained using the virtual environment for the $100$-device scenario.
    \item T-ELLM Original: The original T-ELLM agent, trained for the 20-device scenario, was directly applied without any retraining to test its zero-shot transferability.
    
    \item SAC-Transformer: A baseline using a Transformer architecture with a linear head, trained from scratch on the $100$-device scenario using the SAC algorithm.
\end{itemize}

The results of these experiments are presented in Fig. \ref{scamodel}. As shown in Fig. \ref{sca_energy_eff}, the T-ELLM Retrain model achieves the highest energy efficiency and CWPEM value. Meanwhile, the T-ELLM Original model also performs well, demonstrating strong transferability. The SAC-Transformer baseline is comparatively less efficient. Fig.\ref{sca_acc_per} shows that both T-ELLM Retrain and T-ELLM Original achieve higher convergence accuracy than SAC-Transformer. In other words, the T-ELLM Original model can directly make effective device selections in a much larger environment. Finally, Fig.\ref{sca_energy_per} illustrates that the T-ELLM Retrain is the most energy-efficient. On the contrary, the SAC-Transformer consumes drastically more energy. These results indicate that the T-ELLM framework provides a robust and scalable solution for large-scale FL resource management.

\section{Conclusion and Future Works}
We have presented T-ELLM, a novel tool-aided evolutionary LLM framework for efficient device selection and resource allocation in wireless FL. Notably, T-ELLM has been developed on top of a mathematics-driven decoupling of  LLM-based device selection and tool-based resource allocation. Meanwhile, the combination of the linguistic reasoning capabilities in LLMs and mathematical optimization tools contributes to boosting the generalization capability of decision-making in environmental changes. In addition, T-ELLM takes advantage of a model-based virtual environment to support GRPO-based fine-tuning at minimal communication cost during interactions. Our theoretical analysis has proved the bounded discrepancy between virtual and real environments, which ensures the advantage function learned in the virtual environment maintains a provably small deviation from real-world conditions. Extensive experimental results have demonstrated that T-ELLM can further improve the energy efficiency and exhibit robust adaptability to environmental changes.  Future work will focus on extending the T-ELLM framework to more complicated wireless resource management tasks, such as task offloading \cite{11006640, 10776968}. Key priorities also include investigating its scalability, enhancing its safety against attacks, and validating its performance on real-world hardware. We will also continue to refine its algorithmic complexity to ensure practical deployment.
\label{sec: conclusions}


\bibliographystyle{IEEEtran}
\bibliography{IEEEfull,trans}
\vspace{0.9em}


\end{document}